\documentclass[runningheads]{llncs}
\usepackage{graphicx}
\usepackage{amsmath,amssymb} 
\usepackage{color}
\usepackage{indentfirst}
\usepackage{times}
\usepackage{epsfig}
\usepackage{graphicx}
\usepackage{amsmath}
\usepackage{amssymb}
\usepackage[dvipsnames]{xcolor}
\usepackage{xspace}
\usepackage{multirow}
\usepackage{bbm}
\usepackage{multirow}
\usepackage{microtype}
\usepackage{booktabs}
\usepackage{tabularx}
\usepackage{makecell}
\usepackage{breqn}
\usepackage{float}
\usepackage{adjustbox}

\begin{document}
\newcommand{\zhengqi}[1]{{\textcolor{blue}{[\emph{Zhengqi says}: #1]}}}
\newcommand{\noah}[1]{{\textcolor{orange}{[\emph{Noah says}: #1]}}}

\newcommand{\ICG}{{\sc CGIntrinsics}\xspace}
\newcommand{\ICGShort}{{\sc CGI}\xspace}

\newcommand{\BT}{{\sc BigTime}\xspace}
\newcommand{\BTShort}{{\sc BT}\xspace}

\newcommand{\Images}{\mathcal{I}}
\newcommand{\Reflectances}{\mathcal{R}}
\newcommand{\Shadings}{\mathcal{S}}
\newcommand{\Energy}{\mathcal{E}}
\newcommand{\Mask}{M}
\newcommand{\acronym}{APWLS\xspace}
\newcommand{\diag}{\operatorname{\mathrm{diag}}}

\newcommand{\IHDR}{I_{\mathsf{HDR}}}
\newcommand{\ILDR}{I_{\mathsf{LDR}}}

\newcommand{\Lcg}{\mathcal{L}_{\mathsf{CGI}}}
\newcommand{\Liiw}{\mathcal{L}_{\mathsf{IIW}}}
\newcommand{\lambdaiiw}{\lambda_{\mathsf{IIW}}}
\newcommand{\Lsaw}{\mathcal{L}_{\mathsf{SAW}}}
\newcommand{\lambdasaw}{\lambda_{\mathsf{SAW}}}
\newcommand{\Lsns}{\mathcal{L}_{\mathsf{S/NS}}}
\newcommand{\colorLsns}{{\mathcal{L}_{\mathsf{S/NS}}}}
\newcommand{\lambdasns}{\lambda_{\mathsf{S/NS}}}
\newcommand{\Ls}{\mathcal{L}_{\mathsf{constant-shading}}}
\newcommand{\Lshadow}{\mathcal{L}_{\mathsf{shadow}}}
\newcommand{\Ldata}{\mathcal{L}_{\mathsf{sup}}}
\newcommand{\colorLdata}{{\mathcal{L}_{\mathsf{sup}}}}
\newcommand{\Lsimse}{\mathcal{L}_{\mathsf{siMSE}}}
\newcommand{\colorLsimse}{{\mathcal{L}_{\mathsf{siMSE}}}}
\newcommand{\Lgrad}{\mathcal{L}_{\mathsf{grad}}}
\newcommand{\colorLgrad}{{\mathcal{L}_{\mathsf{grad}}}}
\newcommand{\lambdagrad}{\lambda_{\mathsf{grad}}}
\newcommand{\Lord}{\mathcal{L}_{\mathsf{ord}}}
\newcommand{\colorLord}{{\mathcal{L}_{\mathsf{ord}}}}
\newcommand{\lambdaord}{\lambda_{\mathsf{ord}}}
\newcommand{\Lrec}{\mathcal{L}_{\mathsf{reconstruct}}}
\newcommand{\colorLrec}{{\mathcal{L}_{\mathsf{reconstruct}}}}
\newcommand{\lambdarec}{\lambda_{\mathsf{rec}}}
\newcommand{\Lrsm}{\mathcal{L}_{\mathsf{rsmooth}}}
\newcommand{\colorLrsm}{{\mathcal{L}_{\mathsf{rsmooth}}}}
\newcommand{\lambdars}{\lambda_{\mathsf{rs}}}
\newcommand{\Lssm}{\mathcal{L}_{\mathsf{ssmooth}}}
\newcommand{\colorLssm}{{\mathcal{L}_{\mathsf{ssmooth}}}}
\newcommand{\lambdass}{\lambda_{\mathsf{ss}}}

\newcommand{\wrc}{w_1} 
\newcommand{\wrsm}{w_2} 
\newcommand{\wssm}{w_3} 
\newcommand{\wabs}{w_{\mathsf{abs}}}

\newcommand{\Neighbors}{\mathcal{N}}

\newcommand{\etal}{\textit{et al}.}
\newcommand{\median}{\text{median}}
\newcommand{\vmedian}{v^{\mathsf{med}}}
\newcommand{\vmediannorm}{\overline{v^{\mathsf{med}}}}
\newcommand{\vmediannormpq}{\overline{v^{\mathsf{med}}_{pq}}}
\newcommand{\vmedianweight}{\lambda^{\mathsf{med}}}
\newcommand{\vmediannormweight}{\overline{\lambda^{\mathsf{med}}}}

\newcommand{\fp}{\mathbf{f}_p}
\newcommand{\fq}{\mathbf{f}_q}

\newcommand{\todo}[1]{{{[\emph{TODO}: #1]}}}
\newcommand{\norm}[1]{\left\lVert#1\right\rVert}

\newenvironment{packed_enum}{
\begin{enumerate}
  \setlength{\itemsep}{1pt}
  \setlength{\parskip}{2pt}
  \setlength{\parsep}{0pt}
}{\end{enumerate}}

\newenvironment{packed_item}{
\begin{itemize}
  \setlength{\itemsep}{1pt}
  \setlength{\parskip}{2pt}
  \setlength{\parsep}{0pt}
}{\end{itemize}}

\makeatletter
\def\BState{\State\hskip-\ALG@thistlm}
\makeatother

\title{CGIntrinsics: Better Intrinsic Image Decomposition through Physically-Based Rendering} 

\titlerunning{CGIntrinsics}
%
\author{Zhengqi Li\and
Noah Snavely}

\authorrunning{Zhengqi Li and Noah Snavely}
%

\institute{Department of Computer Science \& Cornell Tech, Cornell University}
\maketitle              
\begin{abstract}
Intrinsic image decomposition is a challenging, long-standing computer
vision problem for which ground truth data is very difficult to
acquire. We explore the use of synthetic data for training CNN-based
intrinsic image decomposition models, then applying these learned
models to real-world images. To that end, we present \ICG, a new,
large-scale dataset of physically-based rendered images of scenes with
full ground truth decompositions. The rendering process we use is
carefully designed to yield high-quality, realistic images, which we
find to be crucial for this problem domain. We also propose a new
end-to-end training method that learns better decompositions by
leveraging \ICG, and optionally IIW and SAW, two recent datasets of
sparse annotations on real-world images. Surprisingly, we find that a
decomposition network trained solely on our synthetic data outperforms
the state-of-the-art on both IIW and SAW, and performance improves
even further when IIW and SAW data is added during training. Our work
demonstrates the suprising effectiveness of carefully-rendered
synthetic data for the intrinsic images task.

\end{abstract}

\section{Introduction}

Intrinsic images is a classic vision problem involving decomposing an
input image $I$ into a product of reflectance (albedo) and shading
images $R\cdot S$. Recent years have seen remarkable progress on this
problem, but it remains challenging due to its ill-posedness. An
attractive proposition has been to replace traditional hand-crafted
priors with learned, CNN-based models. For such learning methods data
is key, but collecting ground truth data for intrinsic images is
extremely difficult, especially for images of real-world scenes.

One way to generate large amounts of training data for intrinsic
images is to render synthetic scenes. However, existing synthetic
datasets are limited to images of single
objects~\cite{janner2017intrinsic,shi2016learning} (e.g., via
ShapeNet~\cite{chang2015shapenet}) or images of CG animation that
utilize simplified, unrealistic illumination (e.g., via
Sintel~\cite{butler2012naturalistic}). An alternative is to collect
ground truth for real images using crowdsourcing, as in the Intrinsic
Images in the Wild (IIW) and Shading Annotations in the Wild (SAW)
datasets~\cite{bell2014intrinsic,kovacs2017shading}. However, the
annotations in such datasets are sparse and difficult to collect
accurately at scale.




Inspired by recent efforts to use synthetic images of scenes as
training data for indoor and outdoor scene
understanding~\cite{richter2016playing,ros2016synthia,gaidon2016virtual,richter2017playing},
we present the first large-scale scene-level intrinsic images dataset
based on high-quality physically-based rendering, which we call \ICG
(\ICGShort). \ICGShort consists of over 20,000 images of indoor
scenes, based on the SUNCG dataset~\cite{song2016semantic}. Our aim
with \ICGShort is to help drive significant progress towards solving
the intrinsic images problem for Internet photos of real-world
scenes. We find that high-quality physically-based rendering is
essential for our task.
While SUNCG provides physically-based scene
renderings~\cite{zhang2017physically}, our experiments show that the
details of how images are rendered are of critical importance,
and certain choices can lead to massive improvements in how well CNNs
trained for intrinsic images on synthetic data generalize to real
data.

We also propose a new partially supervised learning method for
training a CNN
to directly predict reflectance and shading, by combining ground truth
from \ICGShort and sparse annotations from IIW/SAW. Through
evaluations on IIW and SAW, we find that, surprisingly, decomposition
networks trained solely on \ICGShort can achieve state-of-the-art
performance on both datasets. Combined training using both \ICGShort
and IIW/SAW leads to even better performance. Finally, we find that
\ICGShort generalizes better than existing
datasets by evaluating on MIT Intrinsic Images, a very different,
object-centric, dataset.


\begin{figure}[t]
 \centering
 \includegraphics[width=0.9\columnwidth]{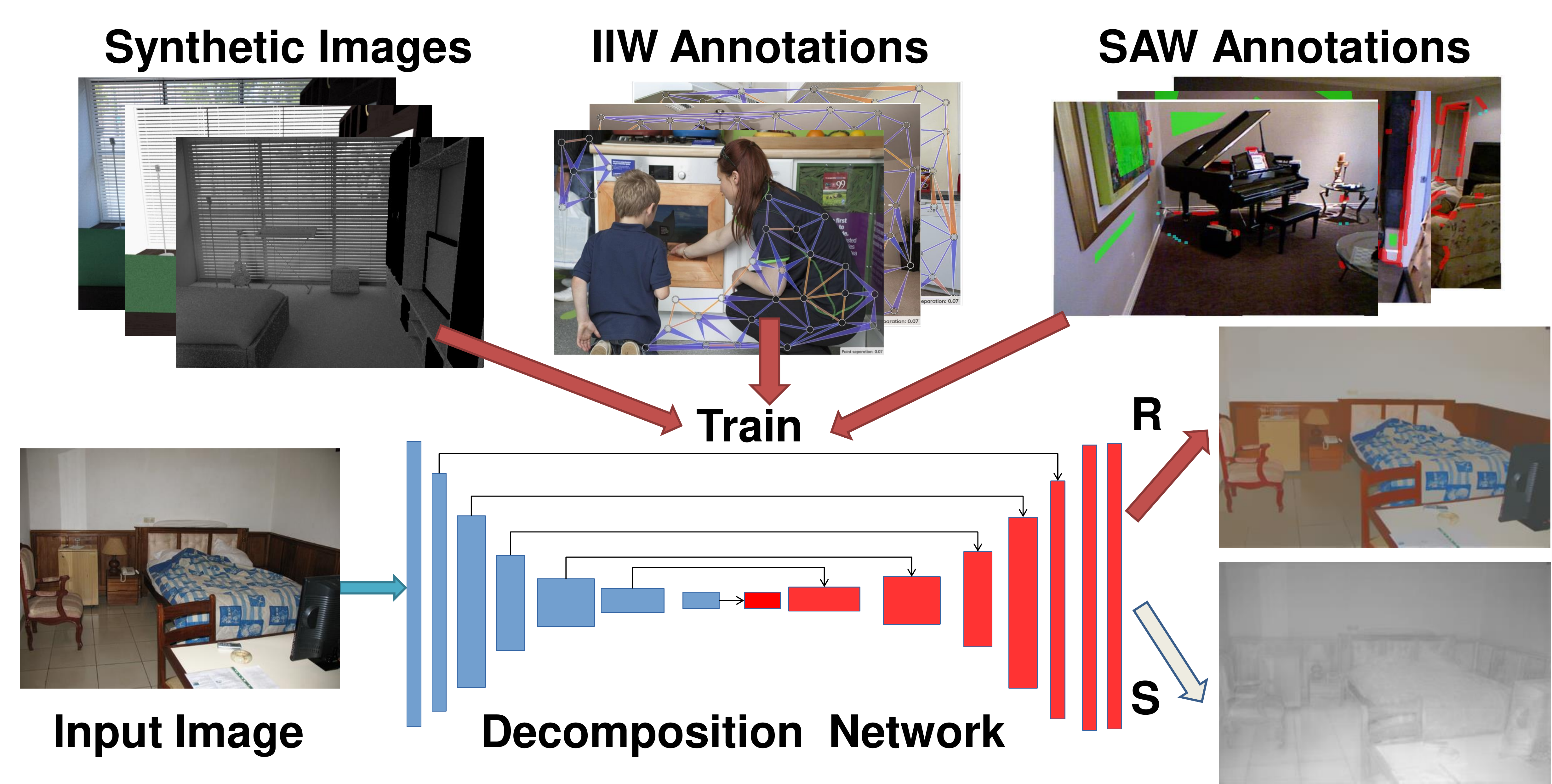} 
 \caption{\textbf{Overview and network architecture.}  Our work
   integrates physically-based rendered images from our \ICG dataset
   and reflectance/shading annotations from IIW and SAW in order to
   train a better intrinsic decomposition
   network. \label{fig:network}}
\end{figure}


\section{Related work}\label{sec:related}

\noindent{\bf Optimization-based methods.}
The classical approach to intrinsic images is to integrate various
priors (smoothness, reflectance sparseness, etc.) into an optimization
framework~\cite{land1971lightness,zhao2012closed,rother2011recovering,shen2011intrinsic,garces2012intrinsic,bell2014intrinsic}. However,
for images of real-world scenes, such hand-crafted prior assumptions
are difficult to craft and are often violated.
Several recent methods seek to improve decomposition quality by
integrating surface normals or depths from RGB-D
cameras~\cite{chen2013simple,barron2013intrinsic,jeon2014intrinsic}
into the optimization process. However, these methods assume depth
maps are available during optimization, preventing them from being
used for a wide range of consumer photos.

\smallskip
\noindent {\bf Learning-based methods.}  Learning methods for
intrinsic images have recently been explored as an alternative to
models with hand-crafted priors, or a way to set the parameters of
such models automatically. Barron and Malik~\cite{barron2015shape}
learn parameters of a model that utilizes sophisticated priors on
reflectance, shape and illumination. This approach works on images of
objects (such as in the MIT dataset), but does not generalize to real
world scenes. More recently, CNN-based methods have been deployed,
including work that regresses directly to the output decomposition
based on various training datasets, such as
Sintel~\cite{narihira2015direct,kim2016unified}, MIT intrinsics and
ShapeNet~\cite{shi2016learning,janner2017intrinsic}. Shu~\etal~\cite{shu2017neural}
also propose a CNN-based method specifically for the domain of facial
images, where ground truth geometry can be obtained through model
fitting. However, as we show in the evaluation section, the networks
trained on such prior datasets perform poorly on images of real-world
scenes.

Two recent datasets are based on images of real-world
scenes. Intrinsic Images in the Wild (IIW)~\cite{bell2014intrinsic}
and Shading Annotations in the Wild (SAW)~\cite{kovacs2017shading}
consist of sparse, crowd-sourced reflectance and shading annotations
on real indoor images. Subsequently, several papers train CNN-based
classifiers on these sparse annotations and use the classifier outputs
as priors to guide
decomposition~\cite{kovacs2017shading,zhou2015learning,zoran2015learning,narihira2015learning}.
However, we find these annotations alone are insufficient to train a
direct regression approach, likely because they are sparse and are
derived from just a few thousand images. Finally, very recent work has
explored the use of time-lapse imagery as training data for intrinsic
images~\cite{Li2018bigtime}, although this provides a very indirect
source of supervision.

\smallskip
\noindent {\bf Synthetic datasets for real scenes.}  Synthetic data
has recently been utilized to improve predictions on real-world images
across a range of problems. For instance,
\cite{richter2016playing,richter2017playing} created a large-scale
dataset and benchmark based on video games for the purpose of
autonomous driving, and
\cite{Beigpour2013IntrinsicIE,bonneel2017intrinsic} use synthetic
imagery to form small benchmarks for intrinsic
images. SUNCG~\cite{zhang2017physically} is a recent, large-scale
synthetic dataset for indoor scene understanding. However, many of the
images in the PBRS database of physically-based renderings derived
from SUNCG have low signal-to-noise ratio (SNR) and non-realistic
sensor properties. We show that higher quality renderings yield much
better training data for intrinsic images.


\section{\ICG Dataset}\label{sec:dataset} 

\begin{figure}[t]
  \centering
    \begin{tabular}{@{}c@{}c@{}c@{}c@{}}
        \includegraphics[width=0.23\columnwidth]{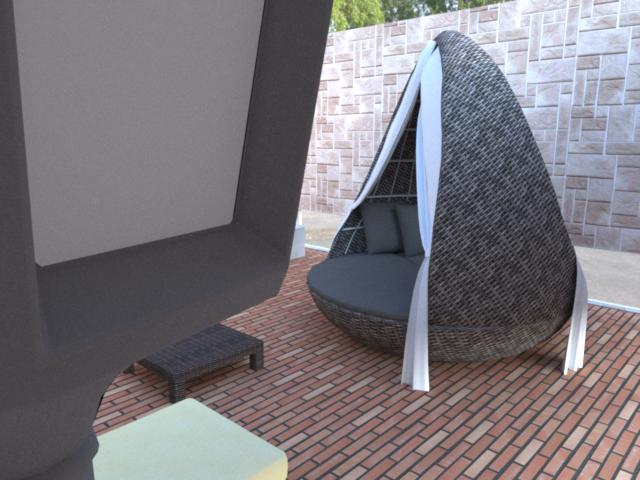}  &
        \includegraphics[width=0.23\columnwidth]{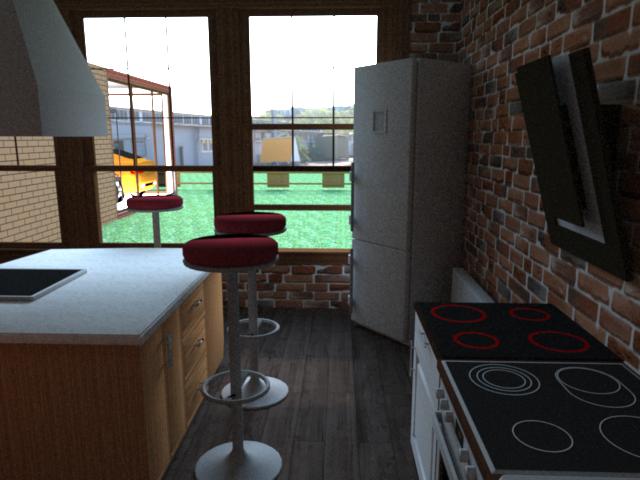}  &
        \includegraphics[width=0.23\columnwidth]{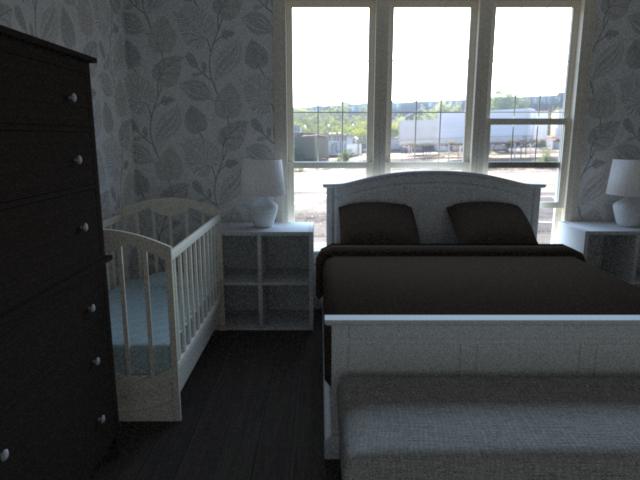}  &
        \includegraphics[width=0.23\columnwidth]{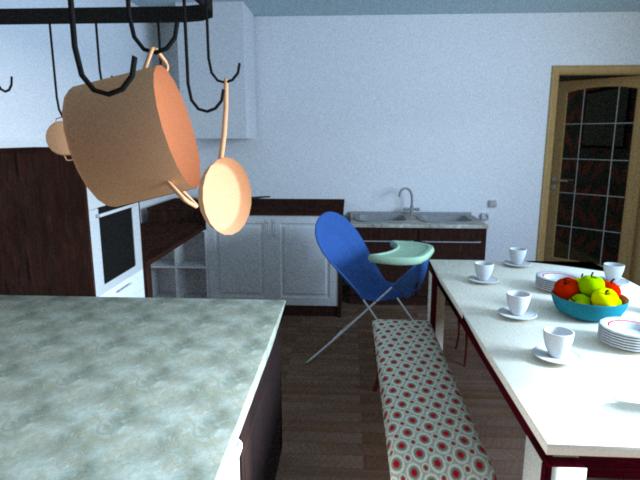}  \\
        \includegraphics[width=0.23\columnwidth]{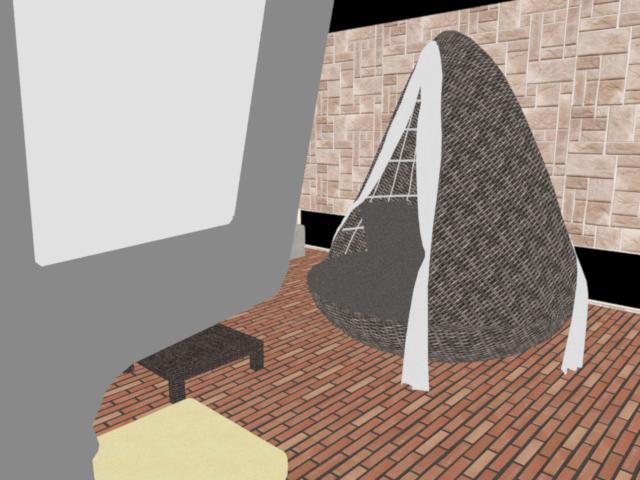}  &
        \includegraphics[width=0.23\columnwidth]{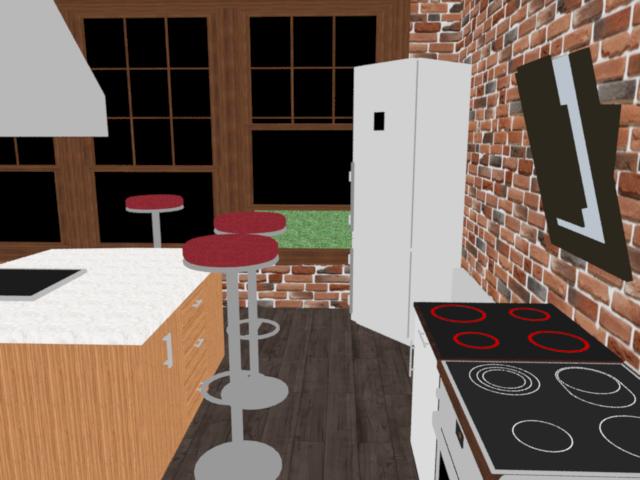}  &
        \includegraphics[width=0.23\columnwidth]{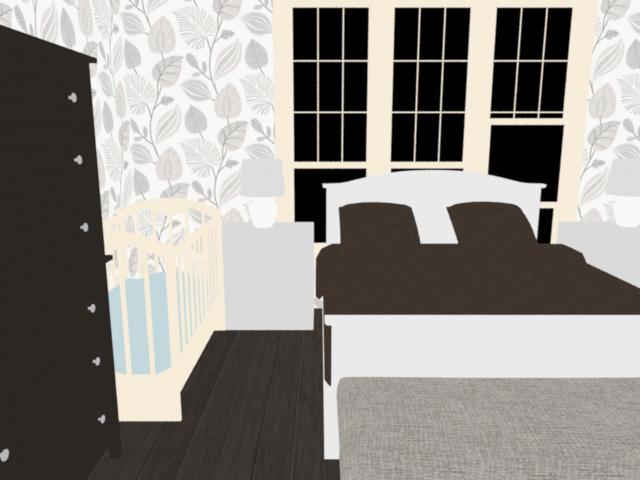}  &
        \includegraphics[width=0.23\columnwidth]{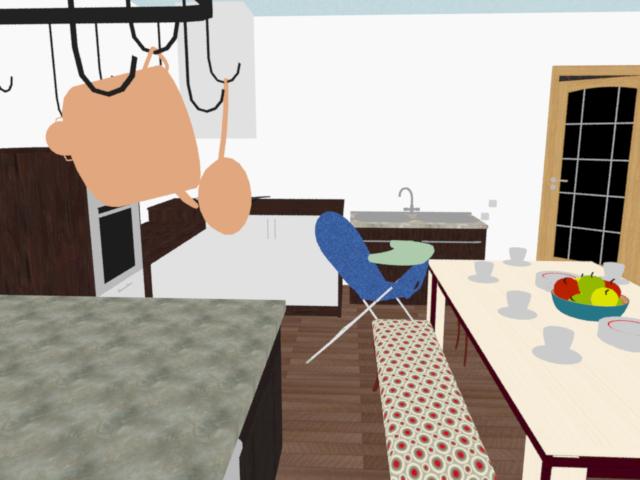}  \\
        \includegraphics[width=0.23\columnwidth]{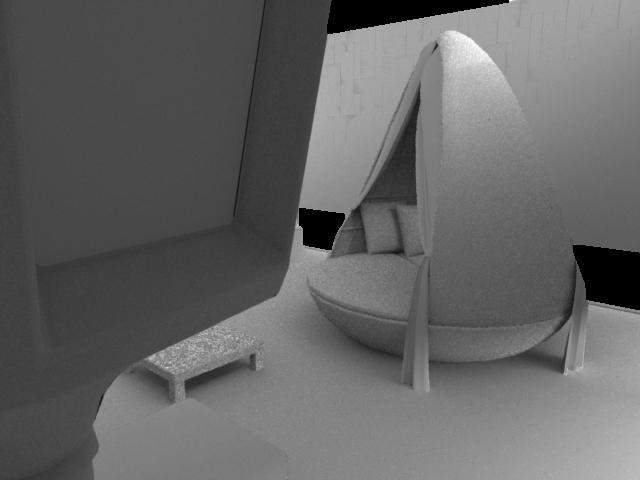}  &
        \includegraphics[width=0.23\columnwidth]{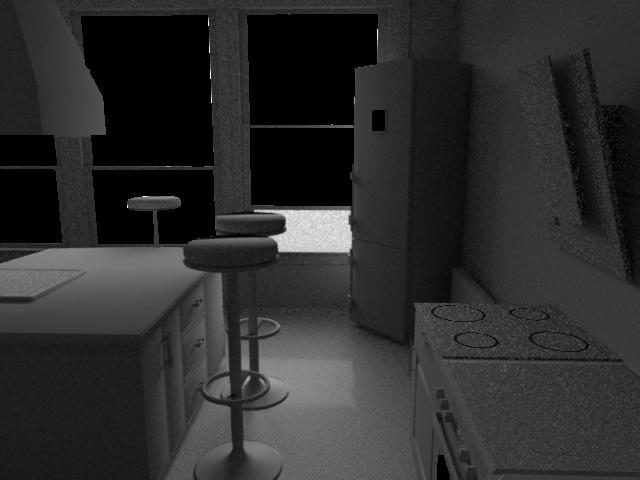}  &
        \includegraphics[width=0.23\columnwidth]{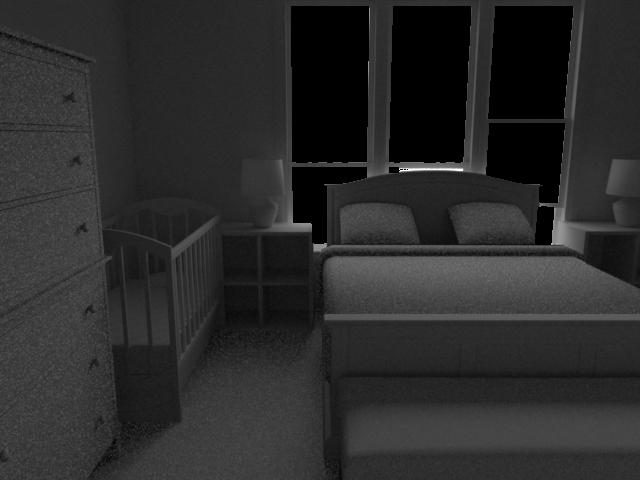}  &
        \includegraphics[width=0.23\columnwidth]{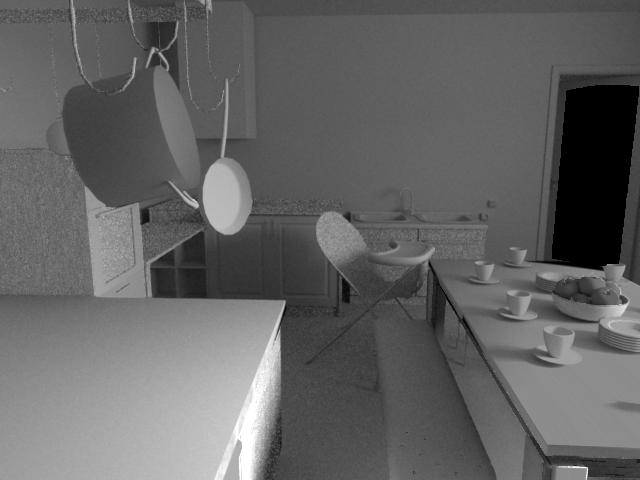}    \\
    \end{tabular}  
  \caption{ \textbf{Visualization of ground truth from our \ICG
      dataset.} Top row: rendered RGB images. Middle: ground truth
    reflectance. Bottom: ground truth shading. Note that light sources
    are masked out when creating the ground truth
    decomposition.}\label{fig:suncg_gt_vis} 
\end{figure}

To create our \ICG (\ICGShort) dataset, we started from the SUNCG
dataset~\cite{song2016semantic}, which contains over 45,000 3D models
of indoor scenes. We first considered the PBRS
dataset of physically-based
renderings of scenes from SUNCG~\cite{zhang2017physically}. For each
scene, PBRS samples cameras from good viewpoints, and uses the
physically-based Mitsuba renderer~\cite{mitsuba} to generate realistic
images under reasonably realistic lighting (including a mix of indoor
and outdoor illumination sources), with global illumination. Using
such an approach, we can also generate ground truth data for intrinsic
images by rendering a standard RGB image $I$, then asking the renderer
to produce a reflectance map $R$ from the same viewpoint, and finally
dividing to get the shading image $S = I / R$. Examples of such ground
truth decompositions are shown in Figure~\ref{fig:suncg_gt_vis}. Note
that we automatically mask out light sources (including illumination
from windows looking outside) when creating the decomposition, and do
not consider those pixels when training the network.

\begin{figure}[t]
  \centering
    \begin{tabular}{@{}c@{}c@{}c@{}c@{}}
        \includegraphics[width=0.23\columnwidth]{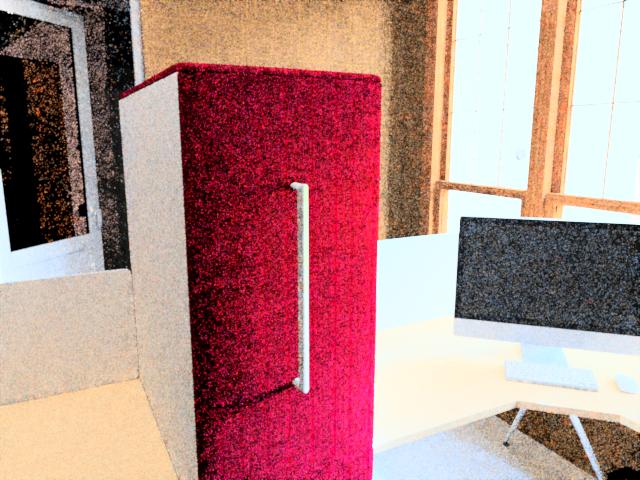} &
        \includegraphics[width=0.23\columnwidth]{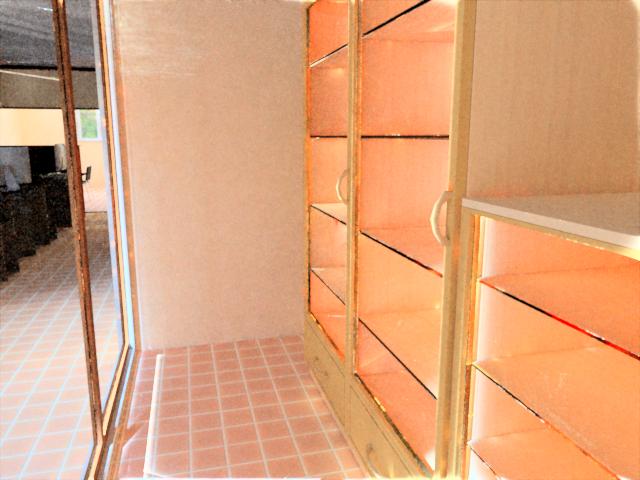}  &        
        \includegraphics[width=0.23\columnwidth]{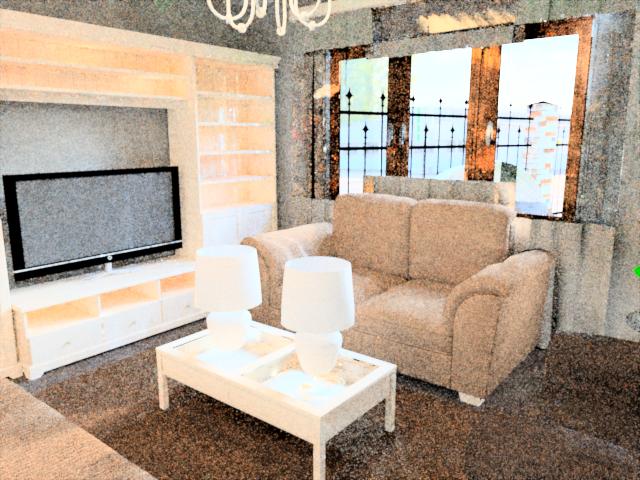}  &
        \includegraphics[width=0.23\columnwidth]{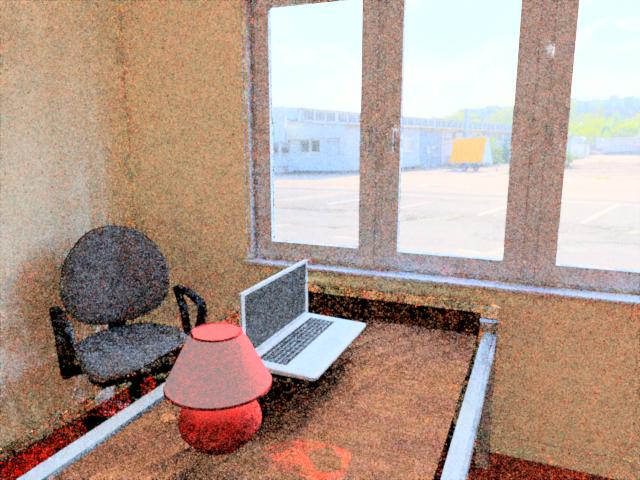} \\
        \includegraphics[width=0.23\columnwidth]{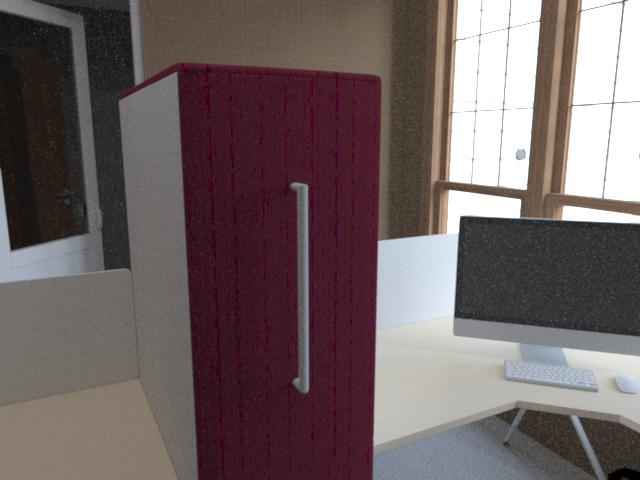}  &
        \includegraphics[width=0.23\columnwidth]{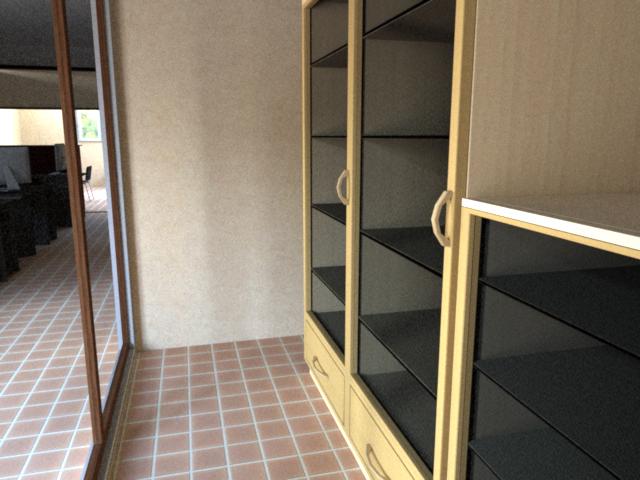}  &
        \includegraphics[width=0.23\columnwidth]{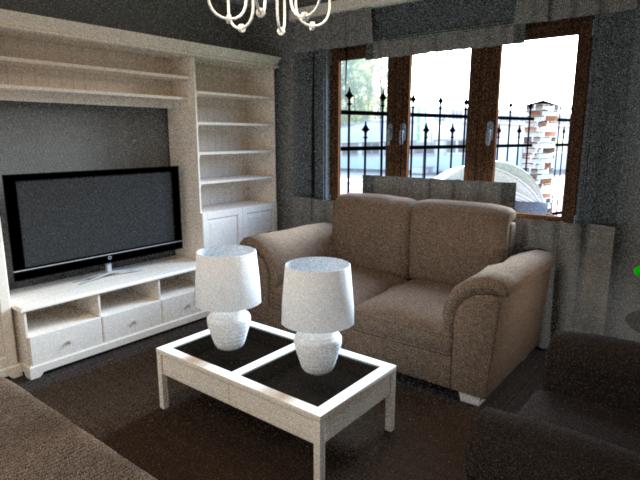}  &
        \includegraphics[width=0.23\columnwidth]{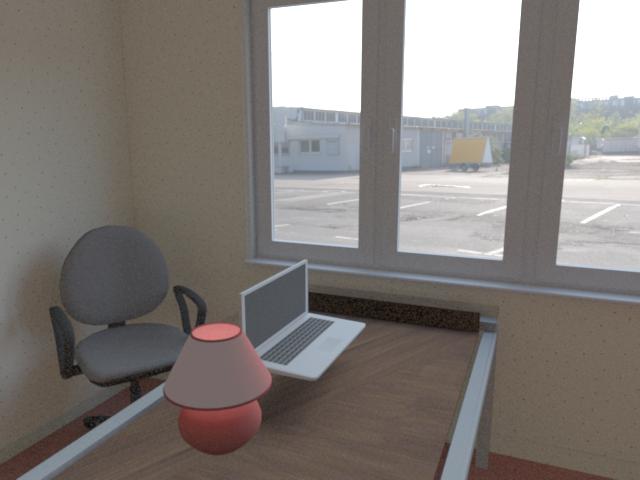}  \\
    \end{tabular}  
  \caption{ \textbf{Visual comparisons between our \ICGShort and the
      original SUNCG dataset.} Top row: images from SUNCG/PBRS. Bottom
    row: images from our \ICGShort dataset. The images in our dataset
    have higher SNR and are more realistic.\label{fig:suncg_visual}} 
\end{figure}

However, we found that the PBRS renderings are not ideal for use in
training real-world intrinsic image decomposition networks. In fact,
certain details in how images are rendered have a dramatic impact on
learning performance:

\smallskip
\noindent {\bf Rendering quality.} Mitsuba and other high-quality
renderers support a range of rendering algorithms, including various
flavors of path tracing methods that sample many light paths for each
output pixel. In PBRS, the authors note that bidirectional path
tracing works well but is very slow, and opt for Metropolis Light
Transport (MLT) with a sample rate of 512 samples per
pixel~\cite{zhang2017physically}. In contrast, for our purposes we
found that bidirectional path tracing (BDPT) with very large numbers
of samples per pixel was the only algorithm that gave consistently
good results for rendering SUNCG images.
Comparisons between selected renderings from PBRS and our new
\ICGShort images are shown in Figure~\ref{fig:suncg_visual}. Note the
significantly decreased noise in our renderings.

This extra quality comes at a cost. We find that using BDPT with 8,192
samples per pixel yields acceptable quality for most images.
This increases the render time per image significantly, from a
reported 31s~\cite{zhang2017physically}, to approximately 30
minutes.\footnote{While high, this is still a fair ways off of
  reported render times for animated films. For instance, each frame
  of Pixar's {\em Monsters University} took a reported 29 hours to
  render~\cite{pixar}.} One reason for the need for large numbers of
samples is that SUNCG scenes are often challenging from a rendering
perspective---the illumination is often indirect, coming from open
doorways or constrained in other ways by geometry. However, rendering
is highly parallelizable, and over the course of about six months we
rendered over ten thousand images on a cluster of about 10 machines.

\smallskip
\noindent {\bf Tone mapping from HDR to LDR.} We found that another
critical factor in image generation is how rendered images are tone
mapped. Renderers like Mitsuba generally produce high dynamic range
(HDR) outputs that encode raw, linear radiance estimates for each
pixel. In contrast, real photos are usually low dynamic range.
The process that takes an HDR input and produces an LDR output is
called {\em tone mapping}, and in real cameras the analogous
operations are the auto-exposure, gamma correction, etc., that yield a
well-exposed, high-contrast photograph. PBRS uses the tone mapping
method of Reinhard~\etal~\cite{reinhard2002photographic}, which is
inspired by photographers such as Ansel Adams, but which can produce
images that are very different in character from those of consumer
cameras. We find that a simpler tone mapping method produces more
natural-looking results. Again, Figure~\ref{fig:suncg_visual} shows
comparisons between PBRS renderings and our own. Note how the color
and illumination features, such as shadows, are better captured in our
renderings (we noticed that shadows often disappear with the Reinhard
tone mapper).

In particular, to tone map a linear HDR radiance image $\IHDR$, we
find the $90^{th}$ percentile intensity value $r_{90}$, then compute
the image $\ILDR = \alpha \IHDR^{\gamma}$, where $\gamma =
\frac{1}{2.2}$ is a standard gamma correction factor, and $\alpha$ is
computed such that $r_{90}$ maps to the value 0.8. The final image is
then clipped to the range $[0,1]$. This mapping ensures that at most
10\% of the image pixels (and usually many fewer) are saturated after
tone mapping, and tends to result in natural-looking
LDR images.

\begin{table*}[tb]
\centering
{\small
\begin{tabular}{lccccc}
  \toprule
  Dataset & Size & Setting & Rendered/Real & Illumination  & GT type  \\
  \midrule
MPI Sintel~\cite{Butler:ECCV:2012} & 890 & Animation & non-PB & spatial-varying & full \\
MIT Intrinsics~\cite{grosse2009ground} & 110 & Object & Real & single global & full \\
ShapeNet~\cite{shi2016learning} & 2M+ & Object & PB & single global & full \\
IIW~\cite{bell2014intrinsic}  &  5230 & Scene & Real & spatial-varying & sparse \\
SAW~\cite{kovacs2017shading}  &  6677 & Scene & Real & spatial-varying & sparse \\
\ICG  &  20,000+ & Scene & PB & spatial-varying & full \\
\bottomrule
\end{tabular} 
\caption{{\bf Comparisons of existing intrinsic image datasets with
    our \ICG dataset.} PB indicates physically-based rendering and
  non-PB indicates non-physically-based rendering. \label{tb:dataset}}
}
\end{table*} 

Using the above rendering approach, we re-rendered $\sim$ 20,000
images from PBRS. We also integrated 152 realistic renderings
from~\cite{bonneel2017intrinsic} into our dataset.
Table~\ref{tb:dataset} compares our \ICGShort dataset to prior
intrinsic image datasets. Sintel is a dataset created for an animated
film, and does not utilize physical-based rendering. Other datasets,
such as ShapeNet and MIT, are object-centered, whereas \ICGShort
focuses on images of indoor scenes, which have more sophisticated
structure and illumination (cast shadows, spatial-varying lighting,
etc). Compared to IIW and SAW, which include images of real scenes,
\ICGShort has full ground truth and and is much more easily collected
at scale.

\section{Learning Cross-Dataset Intrinsics}\label{sec:approach}

In this section, we describe how we use \ICG to jointly train an
intrinsic decomposition network end-to-end, incorporating additional
sparse annotations from IIW and SAW. Our full training loss
considers training data from each dataset:
\begin{align}
  \mathcal{L} = \Lcg + \lambdaiiw \Liiw + \lambdasaw \Lsaw.
\end{align}
where $\Lcg$, $\Liiw$, and $\Lsaw$ are the losses we use for training
from the \ICGShort, IIW, and SAW datasets respectively. The most
direct way to train would be to simply incorporate supervision from
each dataset. In the case of \ICGShort, this supervision consists of
full ground truth. For IIW and SAW, this supervision takes the form of
sparse annotations for each image, as illustrated in
Figure~\ref{fig:network}. However, in addition to supervision, we found that incorporating
smoothness priors into the loss also improves performance.
Our full loss functions thus incorporate a number of terms:
\begin{align}
  \Lcg  =  & \colorLdata + \lambdaord \colorLord  + \lambdarec \colorLrec \\
  \Liiw = & \lambdaord \colorLord + \lambdars \colorLrsm +
  \lambdass \colorLssm + \colorLrec\\
  \Lsaw = & \lambdasns \colorLsns + \lambdars \colorLrsm + \lambdass \colorLssm + \colorLrec
\end{align}

We now describe each term in detail.

%

\subsection{Supervised losses}

\noindent{\bf CGIntrinsics-supervised loss.}  Since the images in our
\ICGShort dataset are equipped with a full ground truth decomposition,
the learning problem for this dataset can be formulated as a direct
regression problem from input image $I$ to output images $R$ and
$S$. However, because the decomposition is only up to an unknown scale
factor, we use a scale-invariant supervised loss, $\Lsimse$ (for
``scale-invariant mean-squared-error''). In addition, we add a
gradient domain multi-scale matching term $\Lgrad$. For each training
image in \ICGShort, our supervised loss is defined as $\Ldata =
\Lsimse + \Lgrad$, where
\begin{gather}
  \Lsimse = \frac{1}{N} \sum_{i=1}^N
  \left( R_i^{*} - c_r R_i \right)^2 + \left( S_i^{*} - c_s S_i
  \right)^2\\
  \Lgrad = \sum_{l=1}^{L} \frac{1}{N_l} \sum_{i=1}^{N_l}
  \norm{\nabla R^{*}_{l,i} - c_r \nabla R_{l,i}}_1 + \norm{\nabla
    S^{*}_{l,i} - c_s \nabla S_{l,i}}_1.
\end{gather}
$R_{l,i}$ ($R_{l,i}^*$) and $S_{l,i}$ ($S_{l,i}^*$) denote reflectance
prediction (resp.\ ground truth) and shading prediction (resp.\ ground
truth) respectively, at pixel $i$ and scale $l$ of an image
pyramid. $N_l$ is the number of valid pixels at scale $l$ and $N =
N_1$ is the number of valid pixels at the original image scale. The
scale factors $c_r$ and $c_s$ are computed via least squares.


In addition to the scale-invariance of $\Lsimse$, another important
aspect is that we compute the MSE in the linear intensity domain, as
opposed to the all-pairs pixel comparisons in the log domain used
in~\cite{narihira2015direct}. In the log domain, pairs of pixels with
large absolute log-difference tend to dominate the loss. As we show in
our evaluation, computing $\Lsimse$ in the linear domain significantly
improves performance.

Finally, the multi-scale gradient matching term $\Lgrad$ encourages
decompositions to be piecewise smooth with sharp discontinuities.

\begin{figure}[t]
  \centering
    \begin{tabular}{@{}c@{}c@{}c@{}c@{}c@{}}
        \includegraphics[width=0.19\textwidth]{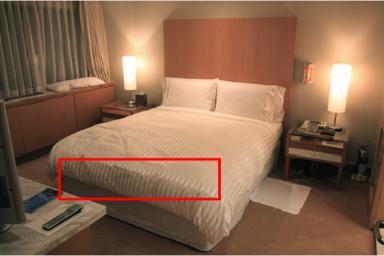}  &
        \includegraphics[width=0.19\textwidth]{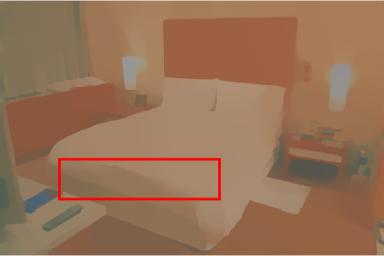}  &
        \includegraphics[width=0.19\textwidth]{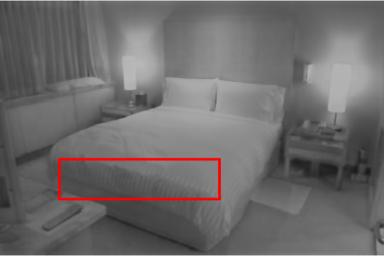}  &
        \includegraphics[width=0.19\textwidth]{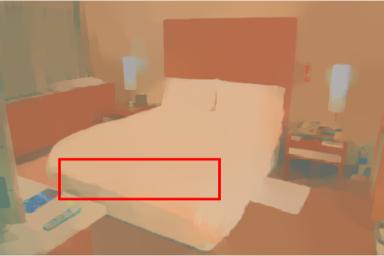}  & 
		\includegraphics[width=0.19\textwidth]{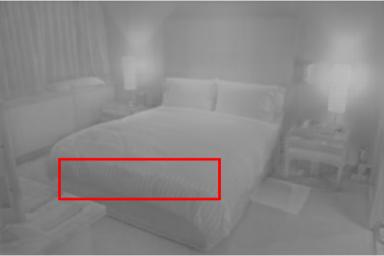}  \\
        \includegraphics[width=0.19\textwidth]{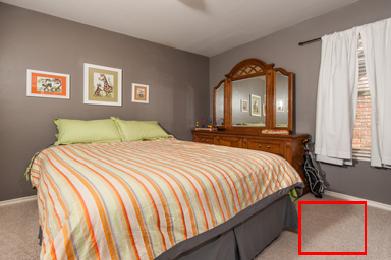}  &
        \includegraphics[width=0.19\textwidth]{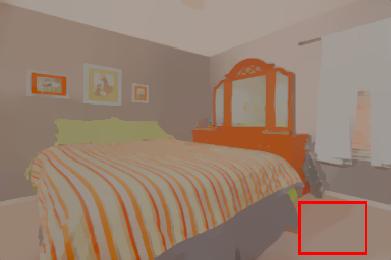}  &
        \includegraphics[width=0.19\textwidth]{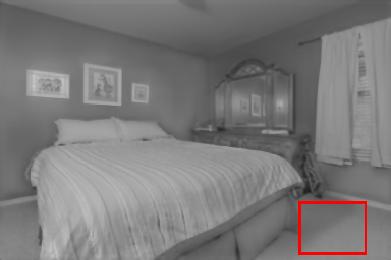} &
        \includegraphics[width=0.19\textwidth]{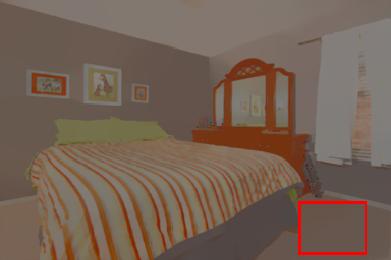}  & 
		\includegraphics[width=0.19\textwidth]{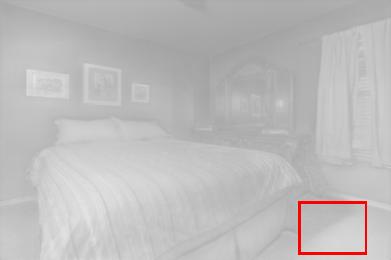}  \\
        {\scriptsize Image} & {\scriptsize \ICGShort ($R$)} & {\scriptsize \ICGShort ($S$)} & {\scriptsize {\ICGShort}+IIW ($R$)}  & {\scriptsize {\ICGShort}+IIW ($S$)} 
    \end{tabular}  
  \caption{ \textbf{Examples of predictions with and without IIW
      training data.} Adding real IIW data can qualitatively improve
    reflectance and shading predictions. Note for instance how the
    quilt highlighted in first row has a more uniform reflectance
    after incorporating IIW data, and similarly for the floor
    highlighted in the second row.} \label{fig:IIW_comparisons1}
\end{figure}

\smallskip
\noindent{\bf Ordinal reflectance loss.}  IIW provides sparse
\emph{ordinal} reflectance judgments between pairs of points (e.g.,
``point $i$ has brighter reflectance than point $j$''). We introduce a
loss based on this ordinal supervision. For a given IIW training image
and predicted reflectance $R$, we accumulate losses for each pair of
annotated pixels $(i,j)$ in that image: $\Lord(R) = \sum_{(i,j)}
e_{i,j}(R)$, where
\begin{align}
  e_{i,j}(R) = 
  \begin{cases}
    w_{i,j} (\log R_i - \log R_j)^2,  &  r_{i,j} = 0 \\ 
    w_{i,j} \left( \max(0, m - \log R_i + \log R_j) \right)^2, & r_{i,j} = +1  \\
    w_{i,j} \left( \max(0, m - \log R_j + \log R_i) \right)^2, & r_{i,j} = -1 \label{eq:ordinal_loss}
  \end{cases}
\end{align} 
and $r_{i,j}$ is the ordinal relation from IIW, indicating whether
point $i$ is darker (-1), $j$ is darker (+1), or they have equal
reflectance (0). $w_{i,j}$ is the confidence of the annotation,
provided by IIW. Example predictions with and without IIW
data are shown in Fig.~\ref{fig:IIW_comparisons1}.

We also found that adding a similar ordinal term derived from
\ICGShort data can improve reflectance predictions. For each image in
\ICGShort, we over-segment it using superpixel
segmentation~\cite{achanta2012slic}. Then in each training iteration,
we randomly choose one pixel from every segmented region, and for each
pair of chosen pixels, we evaluate $\Lord$ similar to
Eq.~\ref{eq:ordinal_loss}, with $w_{i,j} = 1$ and the ordinal relation
derived from the ground truth reflectance.

\smallskip
\noindent{\bf SAW shading loss.}
The SAW dataset provides images containing annotations of smooth (S)
shading regions and non-smooth (NS) shading points, as depicted in
Figure~\ref{fig:network}. These annotations can be further divided
into three types: regions of constant shading, shadow boundaries, and
depth/normal discontinuities.

\begin{figure}[t]
  \centering
    \begin{tabular}{@{}c@{}c@{}c@{}c@{}c@{}}
        \includegraphics[width=0.19\textwidth]{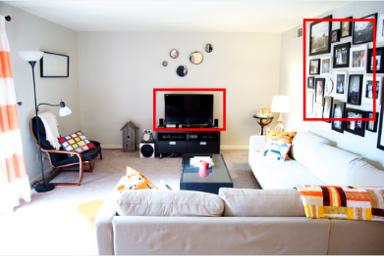}  &
        \includegraphics[width=0.19\textwidth]{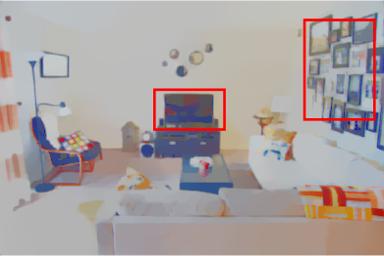}  &
        \includegraphics[width=0.19\textwidth]{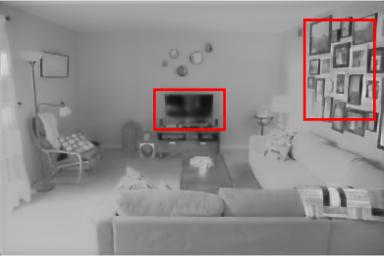}  &
        \includegraphics[width=0.19\textwidth]{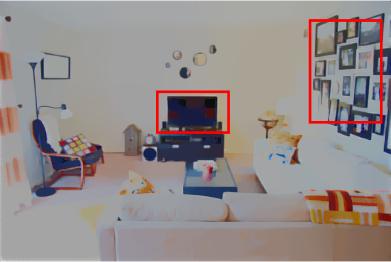} & 
		\includegraphics[width=0.19\textwidth]{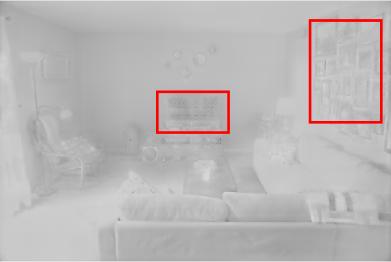} \\
        \includegraphics[width=0.19\textwidth]{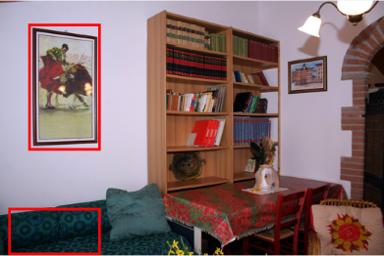}  &
        \includegraphics[width=0.19\textwidth]{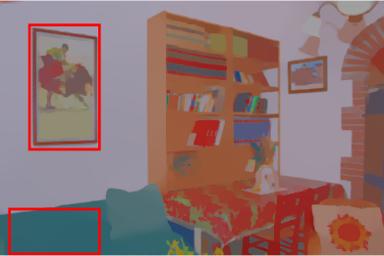}  &
        \includegraphics[width=0.19\textwidth]{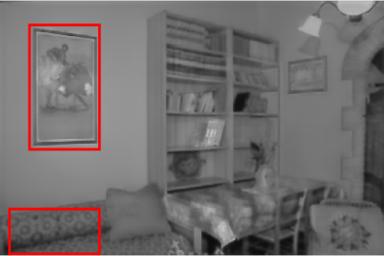}  &
        \includegraphics[width=0.19\textwidth]{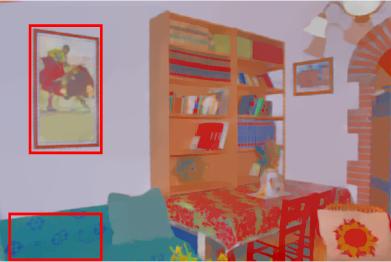}  & 
		\includegraphics[width=0.19\textwidth]{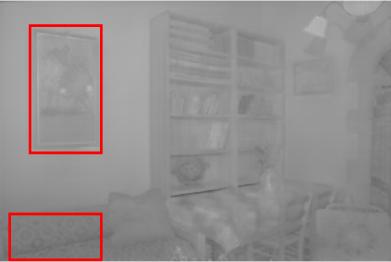}   \\
        {\scriptsize Image} & {\scriptsize \ICGShort ($R$)} & {\scriptsize \ICGShort ($S$)} & {\scriptsize {\ICGShort}+SAW ($R$)}  & {\scriptsize {\ICGShort}+SAW ($S$)} 
    \end{tabular}  
  \caption{ \textbf{Examples of predictions with and without SAW
      training data.} Adding SAW training data can qualitatively
    improve reflectance and shading predictions. Note the pictures/TV
    highlighted in the decompositions in the first row, and the
    improved assignment of texture to the reflectance channel for the
    paintings and sofa in the second row.} \label{fig:SAW_comparisons}
\end{figure}

We integrate all three types of annotations into our supervised SAW
loss $\Lsns$. For each constant shading region (with $N_c$ pixels),
we compute a loss $\Ls$ encouraging the variance of the predicted
shading in the region to be zero:
\begin{align}
  \Ls = \frac{1}{N_c} \sum_{i=1}^{N_c} (\log S_i)^2  - \frac{1}{N_c^2} \left( \sum_{i=1}^{N_c} \log S_i \right)^2.  \label{eq:shading_constant_term}
\end{align}
SAW also provides individual point annotations at cast shadow
boundaries.
As noted in~\cite{kovacs2017shading}, these points are not localized
precisely on shadow boundaries, and so we apply a morphological
dilation with a radius of 5 pixels to the set of marked points before
using them in training. This results in shadow boundary regions.
We find that most shadow boundary annotations lie in regions of
constant reflectance, which implies that for all pair of shading
pixels within a small neighborhood, their log difference should be
approximately equal to the log difference of the image intensity. 
This is equivalent to encouraging the variance of $\log S_i - \log
I_i$ within this small region to be $0$~\cite{eigen2014depth}.
Hence, we define the loss for each shadow boundary region (with
$N_{\mathsf{sd}}$) pixels as:
\begin{align}
  \Lshadow = 
  \frac{1}{N_{\mathsf{sd}}} \sum_{i=1}^{N_{\mathsf{sd}}} (\log S_i - \log I_i )^2  - \frac{1}{N_{\mathsf{sd}}^2} \left( \sum_{i=1}^{N_{\mathsf{sd}}} ( \log S_i -\log I_i )\right)^2  \label{eq:shadow_term}
\end{align}
Finally, SAW provides depth/normal discontinuities, which are also
usually shading discontinuities. However, since we cannot derive the
actual shading change for such discontinuities, we simply mask out
such regions in our shading smoothness term $\Lssm$
(Eq.~\ref{eq:shading_smoothness}), i.e., we do not penalize shading
changes in such regions. As above, we first dilate these annotated
regions before use in training. Examples predictions before/after
adding SAW data into our training are shown in
Fig.~\ref{fig:SAW_comparisons}.

\subsection{Smoothness losses}

To further constrain the decompositions for real images in IIW/SAW,
following classical intrinsic image algorithms we add reflectance
smoothness $\Lrsm$ and shading smoothness $\Lssm$ terms. For
reflectance, we use a multi-scale $\ell_1$ smoothness term to
encourage reflectance predictions to be piecewise constant:
\begin{align}
	\Lrsm = \sum_{l=1}^L \frac{1}{N_l l} \sum_{i=1}^{N_l} \sum_{j \in
          \Neighbors(l,i)} v_{l,i,j} \norm{ \log R_{l,i} - \log R_{l,j} }_1
\end{align}
where $\Neighbors(l,i)$ denotes the 8-connected neighborhood of the
pixel at position $i$ and scale $l$. The reflectance weight $v_{l,i,j}
= \exp\left( - \frac{1}{2} (\mathbf{f}_{l,i} - \mathbf{f}_{l,j})^T
\Sigma^{-1} (\mathbf{f}_{l,i} - \mathbf{f}_{l,j}) \right)$, and the
feature vector $\mathbf{f}_{l,i}$ is defined as $[\ \mathbf{p}_{l,i},
  I_{l,i} , c_{l,i}^1, c_{l,i}^2 \ ]$, where $\mathbf{p}_{l,i}$ and
$I_{l,i}$ are the spatial position and image intensity respectively,
and $c_{l,i}^1$ and $c_{l,i}^2$ are the first two elements of
chromaticity. $\Sigma$ is a covariance matrix defining the distance
between two feature vectors.

We also include a densely-connected $\ell_2$ shading
smoothness term, which can be evaluated in linear time in the number of pixels $N$ using bilateral
embeddings~\cite{barron2015fast,Li2018bigtime}:
\begin{align}
	\Lssm = & \frac{1}{2N} \sum_{i}^N \sum_{j}^N \hat{W}_{i,j} \left( \log S_i - \log S_j \right)^2 
		\approx  \frac{1}{N} \mathbf{s}^\top (I - N_b S_b^\top \bar{B_b} S_b N_b ) \mathbf{s}  \label{eq:shading_smoothness}
\end{align}
where $\hat{W}$ is a bistochastic weight matrix derived from $W$ and
$W_{i,j} = \exp \left( - \frac{1}{2} || \frac{ \mathbf{p}_i
  -\mathbf{p}_j }{\sigma_p} ||_2^2 \right)$.  We refer readers to
\cite{barron2015fast,Li2018bigtime} for a detailed derivation. As
shown in our experiments, adding such smoothness terms to real data
can yield better generalization.

\subsection{Reconstruction loss}
Finally, for each training image in each dataset, we add a loss
expressing the constraint that the reflectance and shading should
reconstruct the original image:
\begin{equation}
  \Lrec = \frac{1}{N} \sum_{i=1}^N \left( I_i - R_i S_i \right)^2.
\end{equation}
  



\subsection{Network architecture}
Our network architecture is illustrated in
Figure~\ref{fig:network}. We use a variant of the ``U-Net''
architecture~\cite{Li2018bigtime,isola2016image}. Our network has one
encoder and two decoders with skip connections. The two decoders
output log reflectance and log shading, respectively. Each layer of
the encoder mainly consists of a $4\times 4$ stride-2 convolutional
layer followed by batch normalization~\cite{ioffe2015batch} and leaky
ReLu~\cite{he2015delving}. For the two decoders, each layer is
composed of a $4\times 4$ deconvolutional layer followed by batch
normalization and ReLu, and a $1\times 1$ convolutional layer is
appended to the final layer of each decoder.

\section{Evaluation}\label{sec:eval}

We conduct experiments on two datasets of real world scenes,
IIW~\cite{bell2014intrinsic} and SAW~\cite{kovacs2017shading} (using
test data unseen during training) and compare our method with several
state-of-the-art intrinsic images algorithms. Additionally, we also
evaluate the generalization of our \ICGShort dataset by evaluating it
on the MIT Intrinsic Images benchmark~\cite{grosse2009ground}.

%



\smallskip
\noindent{\bf Network training details.} We implement our method in
PyTorch~\cite{pytorch}. For all three datasets, we perform data
augmentation through random flips, resizing, and crops. For all
evaluations, we train our network from scratch using the
Adam~\cite{Kingma2014AdamAM} optimizer, with initial learning rate
$0.0005$ and mini-batch size 16. We refer readers to the supplementary
material for the detailed hyperparameter settings.


\subsection{Evaluation on IIW}


\begin{table}[t]
\centering
\setlength\tabcolsep{4pt}
\begin{minipage}{0.48\textwidth}
\centering
\begin{adjustbox}{max width=\textwidth}
\begin{tabular}{llr}\hline
 \toprule
Method & Training set & WHDR \\
\midrule
Retinex-Color~\cite{grosse2009ground} & - & 26.9\% \\
Garces~\etal~\cite{garces2012intrinsic} & - & 24.8\% \\
Zhao~\etal~\cite{zhao2012closed} & - & 23.8\% \\
Bell~\etal~\cite{bell2014intrinsic} & - & 20.6\%  \\
\midrule
Zhou~\etal~\cite{zhou2015learning} & IIW & 19.9\%  \\
\midrule
Bi~\etal~\cite{Bi2015AnLI} & - & 17.7\%  \\
Nestmeyer~\etal~\cite{nestmeyer2017reflectance} & IIW & 19.5\% \\
Nestmeyer~\etal~\cite{nestmeyer2017reflectance}${}^{*}$ & IIW & 17.7\% \\
DI~\cite{narihira2015direct} & Sintel & 37.3\% \\
Shi~\etal~\cite{shi2016learning} & ShapeNet & 59.4\% \\
\bottomrule
\end{tabular}
\end{adjustbox}

\end{minipage}%
\hfill
\begin{minipage}{0.45\textwidth}
\centering
\begin{adjustbox}{max width=\textwidth}
\begin{tabular}{llr}\hline
\toprule
Method & Training set & WHDR \\
\midrule
Ours (log, $\Lsimse$) & \ICGShort & 22.7\% \\
Ours (w/o $\Lgrad$) & \ICGShort & 19.7\% \\
Ours (w/o $\Lord$) & \ICGShort & 19.9\% \\
Ours (w/o $\Lrsm$) & All & 16.1\% \\
Ours & SUNCG & 26.1\% \\
\midrule
Ours${}^{\dag}$ & \ICGShort & 18.4\% \\
Ours & \ICGShort & 17.8\% \\
Ours${}^{*}$ & \ICGShort & 17.1\% \\
Ours & {\ICGShort}+IIW(O) & 17.5\% \\
Ours & {\ICGShort}+IIW(A) & 16.2\% \\
Ours & All  & \textbf{15.5\%} \\
Ours${}^{*}$ & All  & \textbf{14.8\%} \\

\bottomrule
\end{tabular}
\end{adjustbox}
\end{minipage}
\caption{{\bf Numerical results on the IIW test set.} Lower is better
  for WHDR. The ``Training set'' column specifies the training data
  used by each learning-based method: ``-'' indicates an
  optimization-based method. IIW(O) indicates original IIW annotations
  and IIW(A) indicates augmented IIW comparisons. ``All'' indicates
  {\ICGShort}+IIW(A)+SAW. ${}^{\dag}$ indicates network was validated on \ICGShort and others were validated on IIW. ${}^{*}$ indicates CNN predictions are
  post-processed with a guided
  filter~\cite{nestmeyer2017reflectance}. \label{tb:tb_IIW}} 
\end{table}

We follow the train/test split for IIW provided
by~\cite{narihira2015learning}, also used
in~\cite{zhou2015learning}. We also conduct several ablation studies
using different loss configurations. Quantitative comparisons of
Weighted Human Disagreement Rate (WHDR) between our method and other
optimization- and learning-based methods are shown in
Table~\ref{tb:tb_IIW}.

Comparing direct CNN predictions, our \ICGShort-trained model is
significantly better than the best learning-based method
\cite{nestmeyer2017reflectance}, and similar to \cite{Bi2015AnLI},
even though \cite{nestmeyer2017reflectance} was directly trained on
IIW. Additionally, running the post-processing
from~\cite{nestmeyer2017reflectance} on the results of the
\ICGShort-trained model achieves a further performance boost.
Table~\ref{tb:tb_IIW} also shows that models trained on SUNCG (i.e.,
PBRS), Sintel, MIT Intrinsics, or ShapeNet generalize poorly to IIW
likely due to the lower quality of training data (SUNCG/PBRS), or the
larger domain gap with respect to images of real-world scenes,
compared to \ICGShort. The comparison to SUNCG suggests the key
importance of our rendering decisions.

We also evaluate networks trained jointly using \ICGShort and real
imagery from IIW. As in~\cite{zhou2015learning}, we augment the
pairwise IIW judgments by globally exploiting their transitivity and
symmetry.
The right part of Table~\ref{tb:tb_IIW}
demonstrates that including IIW training data leads to further
improvements in performance, as does also including SAW training data.
Table~\ref{tb:tb_IIW} also shows various ablations on variants of our
method, such as evaluating losses in the log domain and removing terms
from the loss functions. Finally, we test a network trained on
\emph{only} IIW/SAW data (and not \ICGShort), or trained on \ICGShort
and fine-tuned on IIW/SAW. Although such a network achieves $\sim$19\%
WHDR, we find that the decompositions are qualitatively
unsatisfactory.  The sparsity of the training data causes these
networks to produce degenerate decompositions, especially for shading
images.

\subsection{Evaluation on SAW}

\begin{table}[t]
\centering
{\small
\begin{tabular}{llrr}
 \toprule
Method & Training set & AP\% (unweighted) &  AP\% (challenge)\\
\midrule
Retinex-Color~\cite{grosse2009ground} & - & 91.93 & 85.26 \\
Garces~\etal~\cite{garces2012intrinsic} & - & 96.89 & 92.39 \\
Zhao~\etal~\cite{zhao2012closed} & - & 97.11 & 89.72 \\
Bell~\etal~\cite{bell2014intrinsic} & - & 97.37 & 92.18\\
\midrule
Zhou~\etal~\cite{zhou2015learning} & IIW & 96.24 & 86.34  \\
Nestmeyer~\etal~\cite{nestmeyer2017reflectance} & IIW & 97.26 & 89.94 \\
Nestmeyer~\etal~\cite{nestmeyer2017reflectance}${}^{*}$ & IIW & 96.85 & 88.64 \\
\midrule
DI~\cite{narihira2015direct} & Sintel+MIT & 95.04 & 86.08\\
Shi~\etal~\cite{shi2016learning} & ShapeNet & 86.62 & 81.30 \\
\midrule
Ours (log, $\Lsimse$) & \ICGShort & 97.73 & 93.03 \\
Ours (w/o $\Lgrad$) & \ICGShort & 98.15 & 93.74 \\
Ours (w/o $\Lssm$) & {\ICGShort}+IIW(A)+SAW & 98.60 & 94.87 \\
Ours & SUNCG & 96.56 & 87.09 \\
\midrule
Ours${}^{\dag}$ & \ICGShort & 98.16 & 93.21 \\
Ours & \ICGShort & 98.39 & 94.05 \\
Ours & {\ICGShort}+IIW(A) & 98.56 & 94.69 \\
Ours & {\ICGShort}+IIW(A)+SAW & \textbf{99.11} & \textbf{97.93} \\
\bottomrule
\end{tabular}
}
\caption{{\bf Quantitative results on the SAW test set.}  Higher is
  better for AP\%. The second column is described in
  Table~\ref{tb:tb_IIW}. The third and fourth columns show performance
  on the unweighted SAW benchmark and our more challenging
  gradient-weighted benchmark,
  respectively. \label{tb:tb_SAW}} 
  \vspace{-0.4cm}
\end{table}

\begin{figure}[!t]
  \centering
    \begin{tabular}{@{}c@{}c@{}}
        \includegraphics[width=0.49\columnwidth]{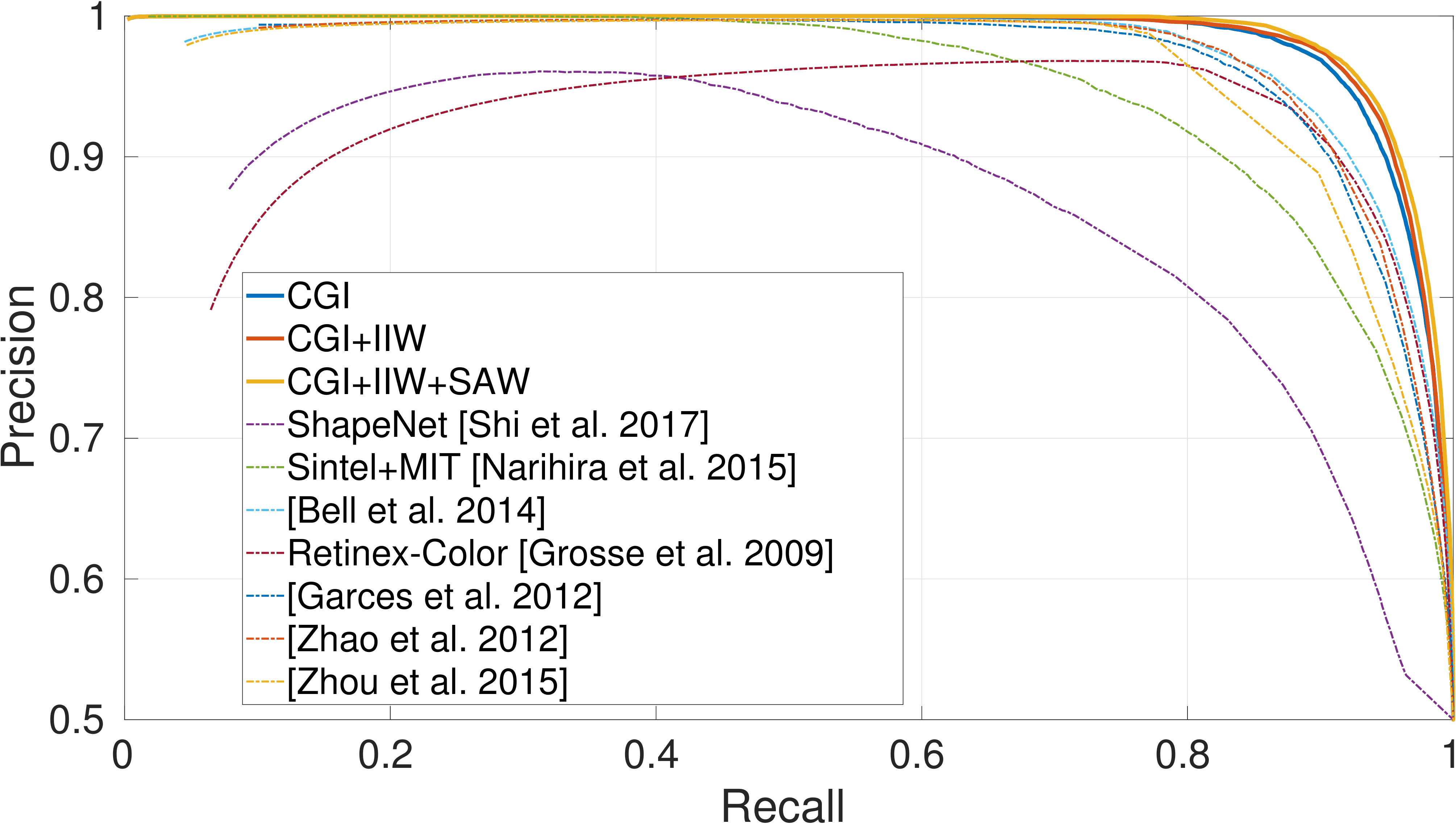}  &
        \includegraphics[width=0.48\columnwidth]{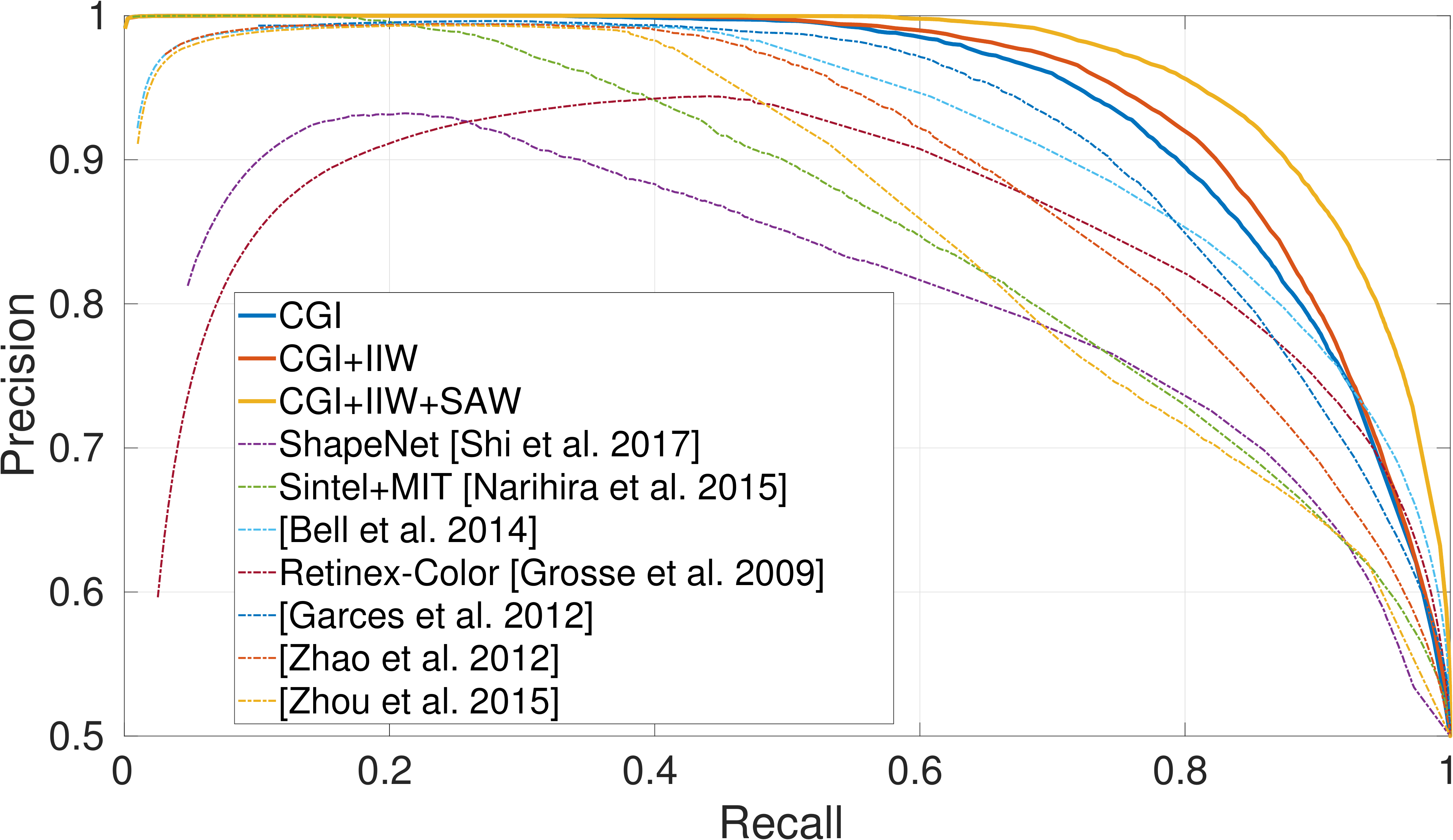}  
    \end{tabular}  
  \caption{ \textbf{Precision-Recall (PR) curve for shading images on
      the SAW test set.} Left: PR curves generated using the
    unweighted SAW error metric of~\cite{Li2018bigtime}. Right: curves
    generated using our more challenging gradient-weighted
    metric.\label{fig:PR_curve_Saw}} 
\end{figure}

To evaluate our shading predictions, we test our models on the
SAW~\cite{kovacs2017shading} test set, utilizing the error metric
introduced in~\cite{Li2018bigtime}. We also propose a new, more
challenging error metric for SAW evaluation.
In particular, we found that many of the constant-shading regions
annotated in SAW also have smooth image intensity (e.g., textureless
walls), making their shading easy to predict. Our proposed metric
downweights such regions as follows. For each annotated region of
constant shading, we compute the average image gradient magnitude over
the region. During evaluation, when we add the pixels belonging to a
region of constant shading into the confusion matrices, we multiply
the number of pixels by this average gradient.  This proposed metric
leads to more distinguishable performance differences between methods,
because regions with rich textures will contribute more to the error
compared to the unweighted metric.

Figure~\ref{fig:PR_curve_Saw} and Table~\ref{tb:tb_SAW} show
precision-recall (PR) curves and average precision (AP) on the SAW
test set with both unweighted~\cite{Li2018bigtime} and our proposed
challenge error metrics. As with IIW, networks trained solely on our
\ICGShort data can achieve state-of-the-art performance, even
\emph{without} using SAW training data. Adding real IIW data improves
the AP in term of both error metrics. Finally, the last column of
Table~\ref{tb:tb_SAW} shows that integrating SAW training data can
significantly improve the performance on shading predictions,
suggesting the effectiveness of our proposed losses for SAW sparse
annotations.

Note that the previous state-of-the-art algorithms on IIW (e.g.,
Zhou~\etal~\cite{zhou2015learning} and
Nestmeyer~\etal~\cite{nestmeyer2017reflectance}) tend to overfit to
reflectance, hurting the accuracy of shading predictions.
This is especially evident in terms of our proposed challenge error
metric. In contrast, our method achieves state-of-the-art results on
\emph{both} reflectance and shading predictions, in terms of all error
metrics. Note that models trained on the original SUNCG, Sintel, MIT
intrinsics or ShapeNet datasets perform poorly on the SAW test set,
indicating the much improved generalization to real scenes of our
\ICGShort dataset.

\begin{figure}[t]
  \centering
    \begin{tabular}{@{}c@{}c@{}c@{}c@{}c@{}c@{}c@{}}
        \includegraphics[width=0.14\textwidth]{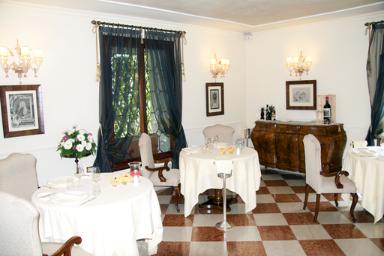}  &
        \includegraphics[width=0.14\textwidth]{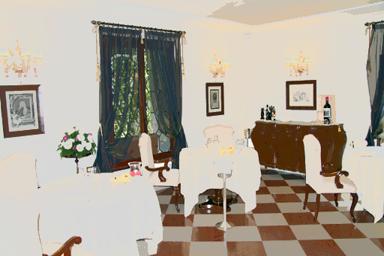}  &
        \includegraphics[width=0.14\textwidth]{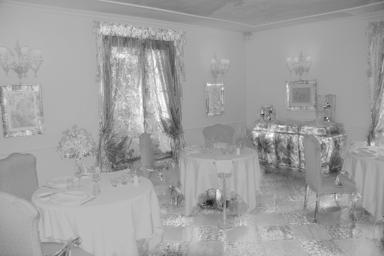}  &
        \includegraphics[width=0.14\textwidth]{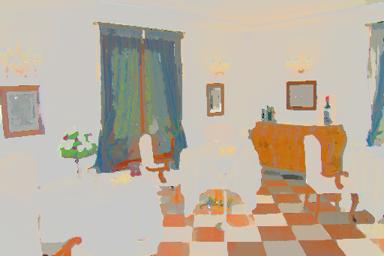}  & 
		\includegraphics[width=0.14\textwidth]{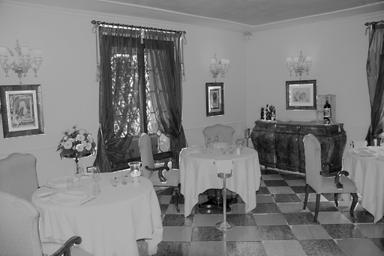}  &         
        \includegraphics[width=0.14\textwidth]{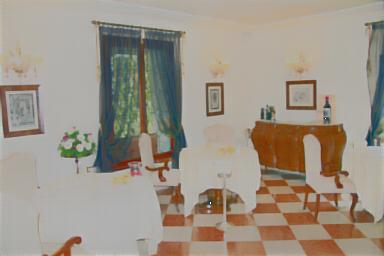}  & 
        \includegraphics[width=0.14\textwidth]{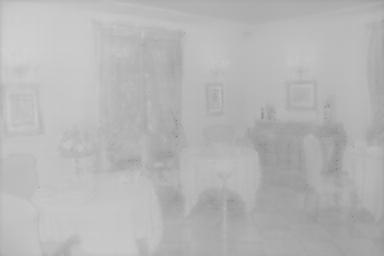}   \\  
        \includegraphics[width=0.14\textwidth]{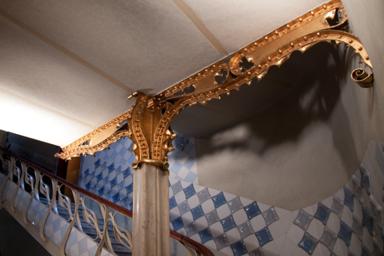}  &
        \includegraphics[width=0.14\textwidth]{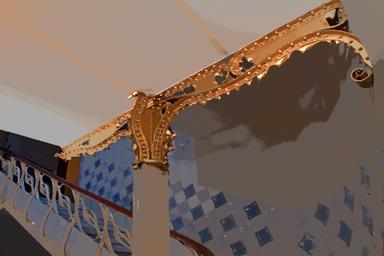}  &
        \includegraphics[width=0.14\textwidth]{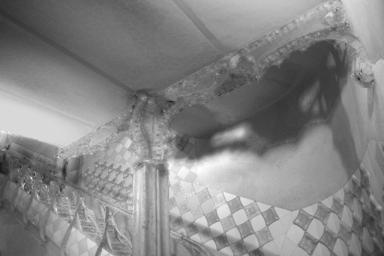} &
        \includegraphics[width=0.14\textwidth]{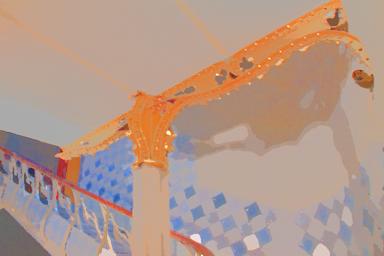} & 
		\includegraphics[width=0.14\textwidth]{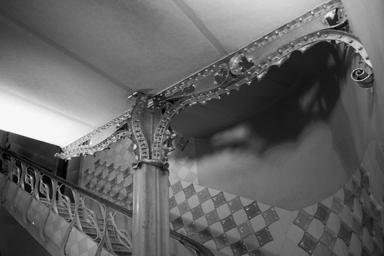}  &         
        \includegraphics[width=0.14\textwidth]{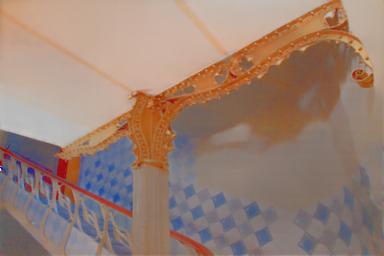}  & 
        \includegraphics[width=0.14\textwidth]{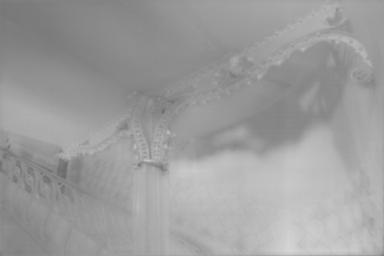}  \\  
        \includegraphics[width=0.14\textwidth]{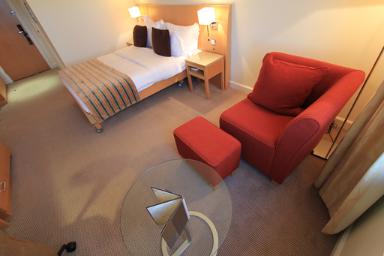} &
        \includegraphics[width=0.14\textwidth]{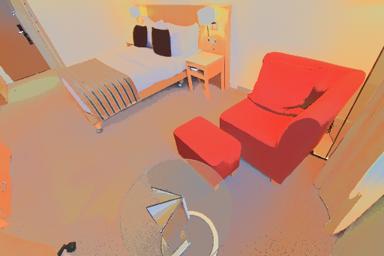} &
        \includegraphics[width=0.14\textwidth]{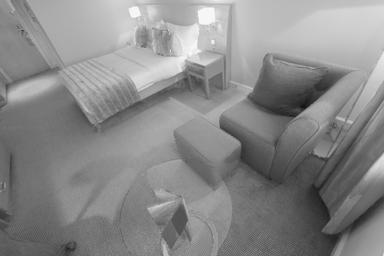} &
        \includegraphics[width=0.14\textwidth]{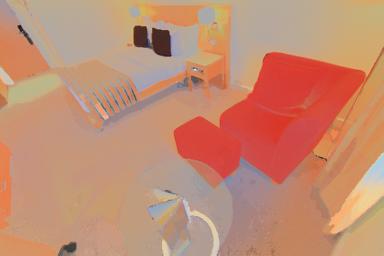} & 
		\includegraphics[width=0.14\textwidth]{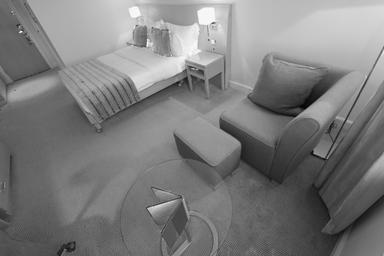} &         
        \includegraphics[width=0.14\textwidth]{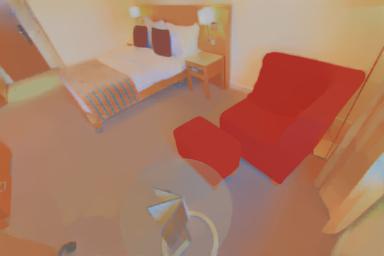}  & 
        \includegraphics[width=0.14\textwidth]{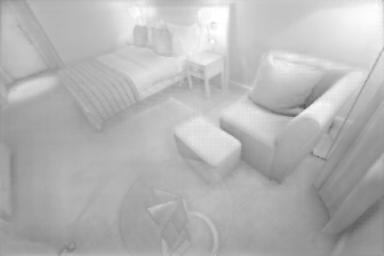}  \\          
        \includegraphics[width=0.14\textwidth]{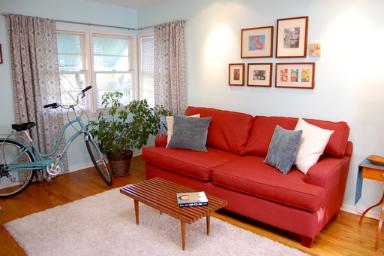}  &
        \includegraphics[width=0.14\textwidth]{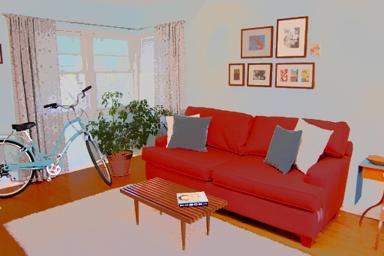}  &
        \includegraphics[width=0.14\textwidth]{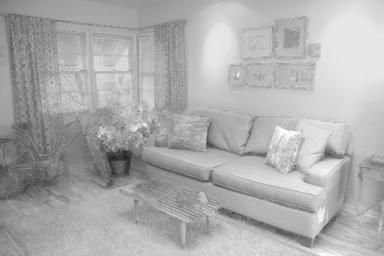}  &
        \includegraphics[width=0.14\textwidth]{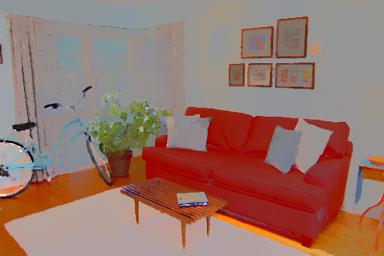}  & 
		\includegraphics[width=0.14\textwidth]{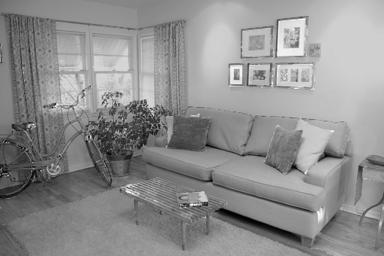} &         
        \includegraphics[width=0.14\textwidth]{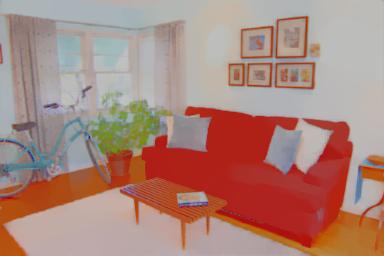}  & 
        \includegraphics[width=0.14\textwidth]{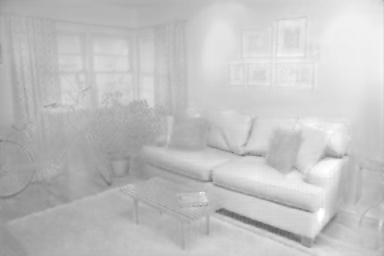}   \\     
        \includegraphics[width=0.14\textwidth]{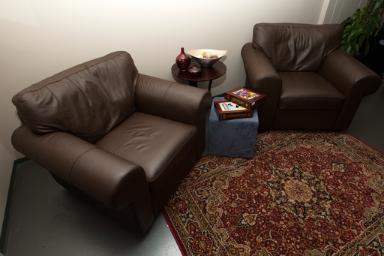}  &
        \includegraphics[width=0.14\textwidth]{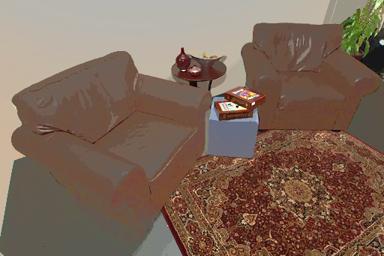}  &
        \includegraphics[width=0.14\textwidth]{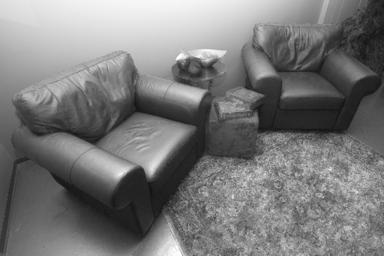}  &
        \includegraphics[width=0.14\textwidth]{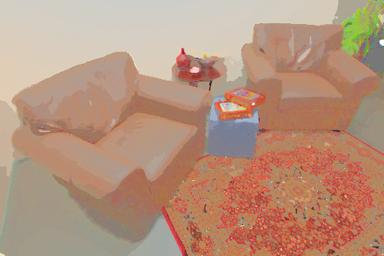} & 
        \includegraphics[width=0.14\textwidth]{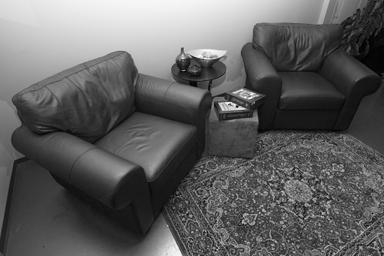} &         
        \includegraphics[width=0.14\textwidth]{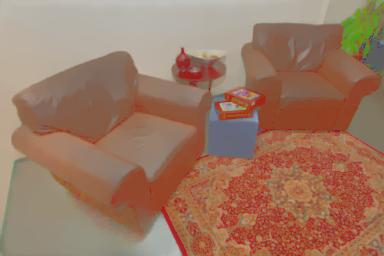} & 
        \includegraphics[width=0.14\textwidth]{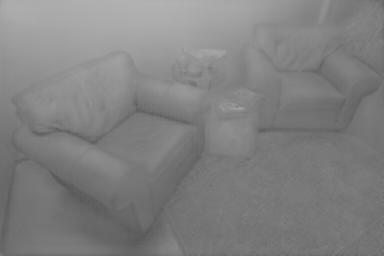} \\
        \includegraphics[width=0.14\textwidth]{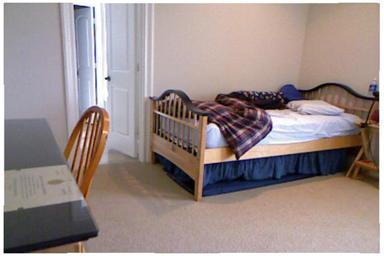}  &
        \includegraphics[width=0.14\textwidth]{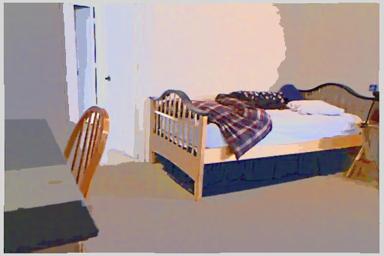}  &
        \includegraphics[width=0.14\textwidth]{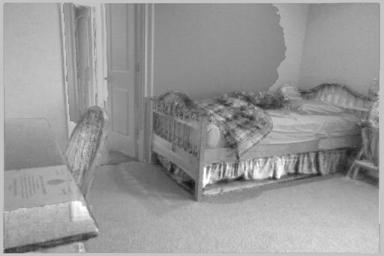}  &
        \includegraphics[width=0.14\textwidth]{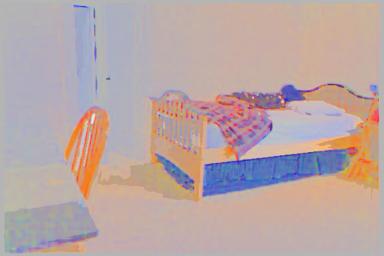} & 
		\includegraphics[width=0.14\textwidth]{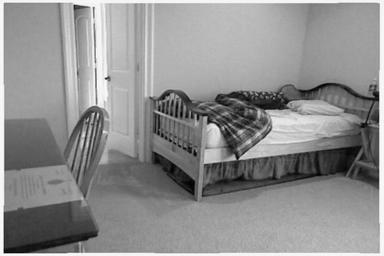} &         
        \includegraphics[width=0.14\textwidth]{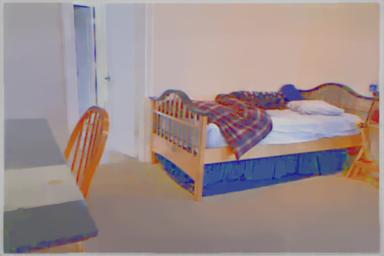} & 
        \includegraphics[width=0.14\textwidth]{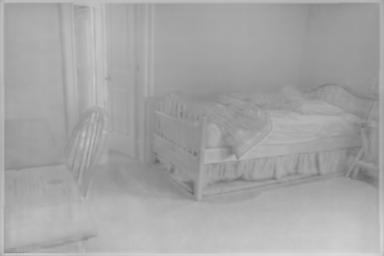} \\
        \includegraphics[width=0.14\textwidth]{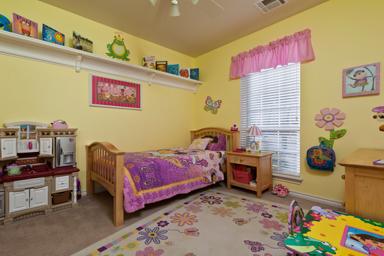}  &
        \includegraphics[width=0.14\textwidth]{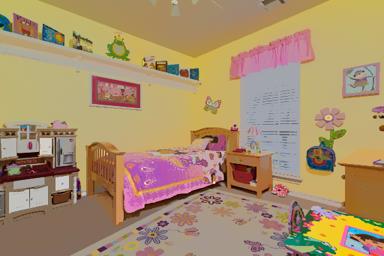}  &
        \includegraphics[width=0.14\textwidth]{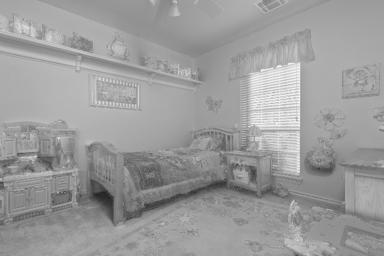}  &
        \includegraphics[width=0.14\textwidth]{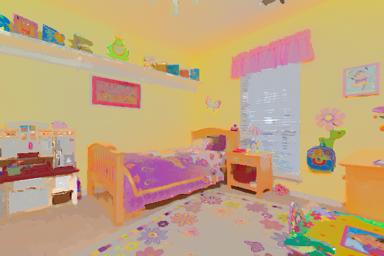} & 
		\includegraphics[width=0.14\textwidth]{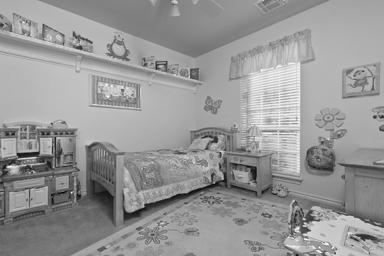}  &         
        \includegraphics[width=0.14\textwidth]{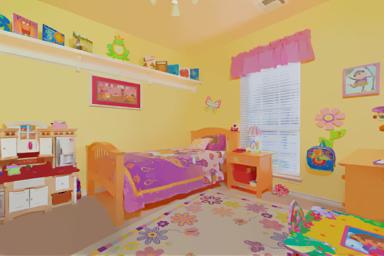}  & 
        \includegraphics[width=0.14\textwidth]{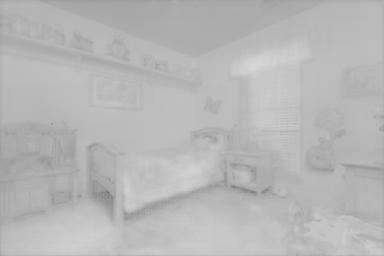}  \\                       
        \includegraphics[width=0.14\textwidth]{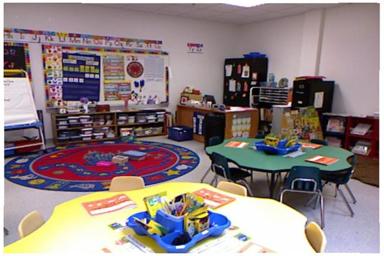}  &
        \includegraphics[width=0.14\textwidth]{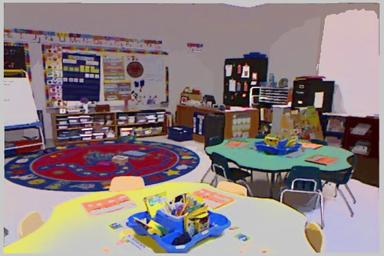}  &
        \includegraphics[width=0.14\textwidth]{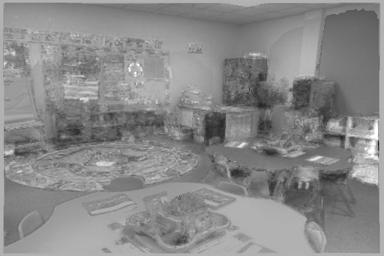}  &
        \includegraphics[width=0.14\textwidth]{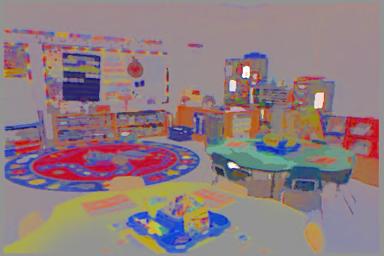} & 
		\includegraphics[width=0.14\textwidth]{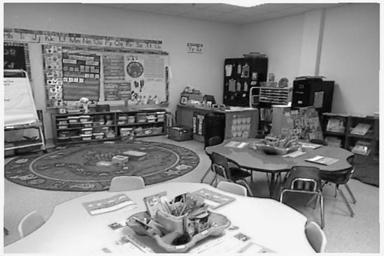}  &         
        \includegraphics[width=0.14\textwidth]{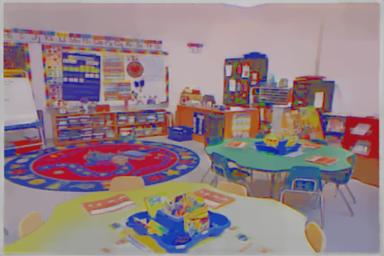}  & 
        \includegraphics[width=0.14\textwidth]{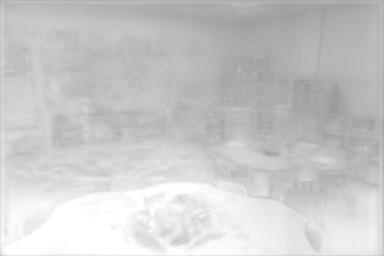}  \\    
        {\scriptsize Image} & {\scriptsize Bell~\etal ($R$)} & {\scriptsize Bell~\etal ($S$)} & {\scriptsize Zhou~\etal ($R$)}  & {\scriptsize Zhou~\etal ($S$)}  & {\scriptsize Ours ($R$)} & {\scriptsize Ours ($S$)}
    \end{tabular}  
  \caption{ \textbf{Qualitative comparisons on the IIW/SAW test sets.}
    Our predictions show significant improvements compared to
    state-of-the-art algorithms (Bell~\etal ~\cite{bell2014intrinsic}
    and Zhou~\etal~\cite{zhou2015learning}). In particular, our
    predicted shading channels include significantly less surface
    texture in several challenging
    settings.} \label{fig:visual_compare} 
\end{figure}

\smallskip
\noindent{\bf Qualitative results on IIW/SAW.}
Figure~\ref{fig:visual_compare} shows qualitative comparisons between
our network trained on all three datasets, and two other
state-of-the-art intrinsic images algorithms
(Bell~\etal~\cite{bell2014intrinsic} and
Zhou~\etal~\cite{zhou2015learning}), on images from the IIW/SAW test
sets. In general, our decompositions show significant improvements. In
particular, our network is better at avoiding attributing surface
texture to the shading channel (for instance, the checkerboard
patterns evident in the first two rows, and the complex textures in
the last four rows) while still predicting accurate reflectance (such
as the mini-sofa in the images of third row). In contrast, the other
two methods often fail to handle such difficult settings. In
particular, \cite{zhou2015learning} tends to overfit to reflectance
predictions,
and their shading estimates strongly resemble the original image
intensity. However, our method still makes mistakes, such as the
non-uniform reflectance prediction for the chair in the fifth row, as
well as residual textures and shadows in the shading and reflectance channels.

\subsection{Evaluation on MIT intrinsic images}

\begin{table}[t]
\centering
{\small
\begin{tabular}{llcccccccc}
\toprule
&  & \multicolumn{2}{c}{MSE} & \phantom{abc} &
\multicolumn{2}{c}{LMSE} & \phantom{abc} & \multicolumn{2}{c}{DSSIM} \\ 
\cmidrule{3-4} \cmidrule{6-7} \cmidrule{9-10}
Method & Training set & refl. & shading && refl. & shading && refl. & shading  \\
\midrule
SIRFS~\cite{barron2015shape} & MIT & \textbf{0.0147} & \textbf{0.0083} && 0.0416 & \textbf{0.0168} && \textbf{0.1238} & \textbf{0.0985} \\
DI~\cite{narihira2015direct} & Sintel+MIT & 0.0277 & 0.0154 && 0.0585 & 0.0295 && 0.1526 & 0.1328 \\
Shi~\etal~\cite{shi2016learning} & ShapeNet & 0.0468 & 0.0194 && 0.0752 & 0.0318 && 0.1825 & 0.1667  \\
Shi~\etal~\cite{shi2016learning}${}^{\star}$ & ShapeNet+MIT & 0.0278 & 0.0126 && 0.0503 & 0.0240 && 0.1465 & 0.1200  \\
Ours & \ICGShort & 0.0221 & 0.0186 && 0.0349 & 0.0259 && 0.1739 & 0.1652  \\
${\text{Ours}}^{\star}$ & \ICGShort+MIT & 0.0167 & 0.0127 && \textbf{0.0319} & 0.0211 && 0.1287 & 0.1376 \\
\bottomrule
\end{tabular} 
\caption{{\bf Quantitative Results on MIT intrinsics testset. } For
  all error metrics, lower is better. The second column shows the
  dataset used for training. ${}^{\star}$ indicates models fine-tuned
  on MIT. \label{tb:tb_MIT}}
}
\end{table}

\begin{figure}[!t]
  \small
  \centering
    \begin{tabular}{@{}c@{}c@{}c@{}c@{}c@{}c@{}c@{}c@{}}
        \includegraphics[width=0.11\textwidth]{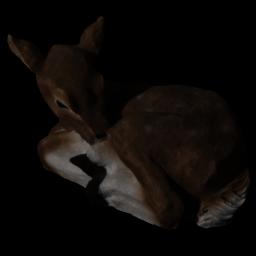}  &
        \includegraphics[width=0.11\textwidth]{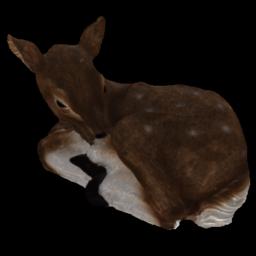} &
        \includegraphics[width=0.11\textwidth]{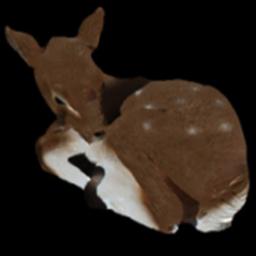}  &
        \includegraphics[width=0.11\textwidth]{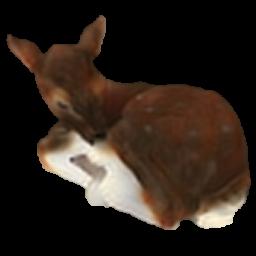}  & 
        \includegraphics[width=0.11\textwidth]{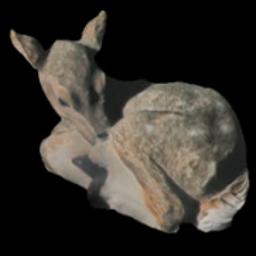}  &         
		\includegraphics[width=0.11\textwidth]{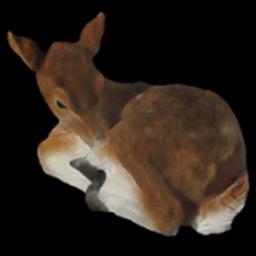}  &
		\includegraphics[width=0.11\textwidth]{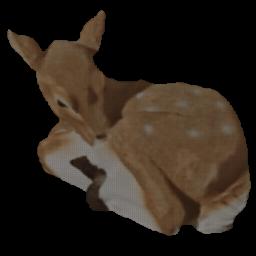}  & 
		\includegraphics[width=0.11\textwidth]{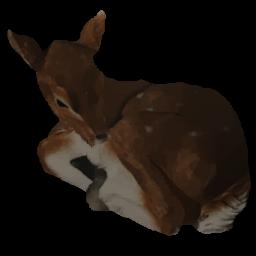}  \\
        \includegraphics[width=0.11\textwidth]{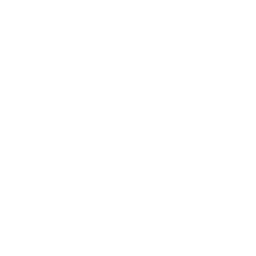}  &
        \includegraphics[width=0.11\textwidth]{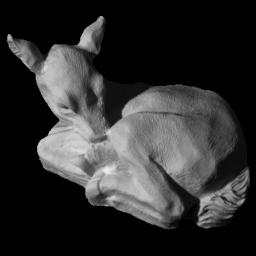} &
        \includegraphics[width=0.11\textwidth]{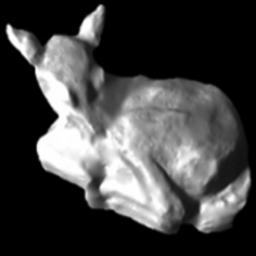}  &
        \includegraphics[width=0.11\textwidth]{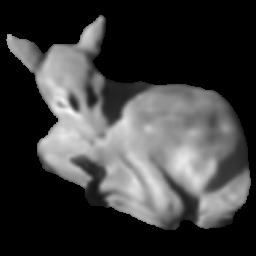}  & 
        \includegraphics[width=0.11\textwidth]{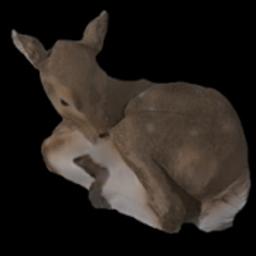}  &          
		\includegraphics[width=0.11\textwidth]{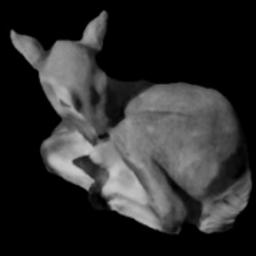}  &
		\includegraphics[width=0.11\textwidth]{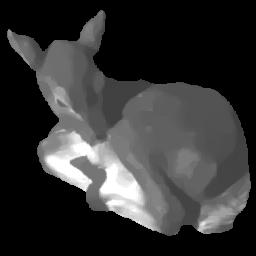}   &
		\includegraphics[width=0.11\textwidth]{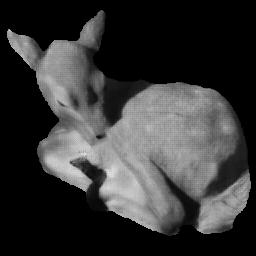}  \\	
        \includegraphics[width=0.11\textwidth]{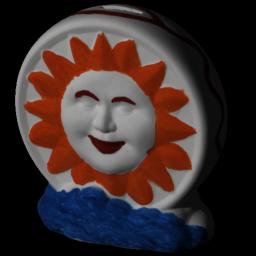}  &
        \includegraphics[width=0.11\textwidth]{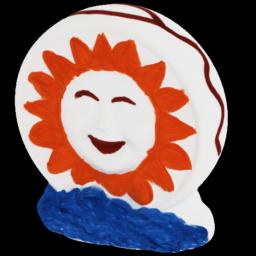}  &
        \includegraphics[width=0.11\textwidth]{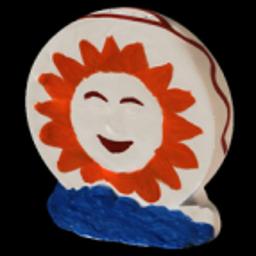}  &
        \includegraphics[width=0.11\textwidth]{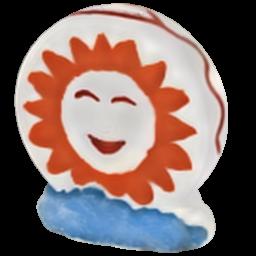}   & 
        \includegraphics[width=0.11\textwidth]{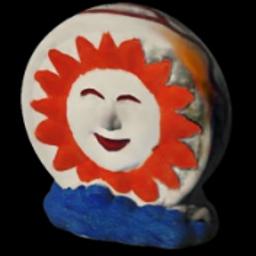}  &          
		\includegraphics[width=0.11\textwidth]{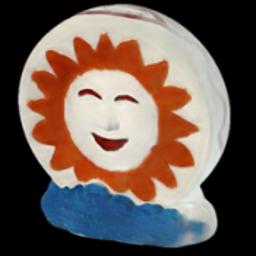}  &
		\includegraphics[width=0.11\textwidth]{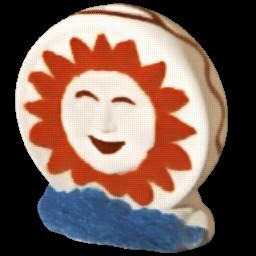}  & 
		\includegraphics[width=0.11\textwidth]{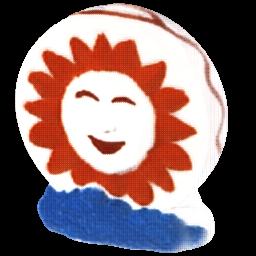}  \\		
        \includegraphics[width=0.11\textwidth]{figs/MIT/ones.png}  &
        \includegraphics[width=0.11\textwidth]{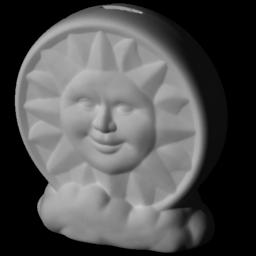}  &
        \includegraphics[width=0.11\textwidth]{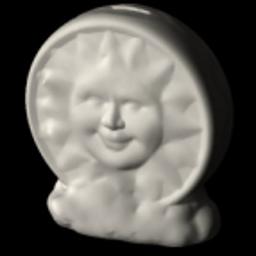}  &
        \includegraphics[width=0.11\textwidth]{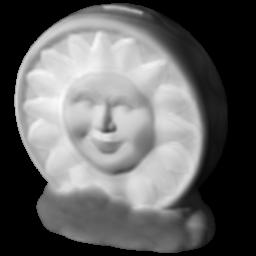}  & 
        \includegraphics[width=0.11\textwidth]{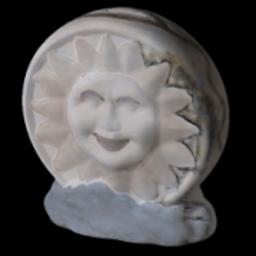}  &          
		\includegraphics[width=0.11\textwidth]{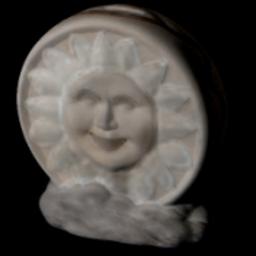}  &
		\includegraphics[width=0.11\textwidth]{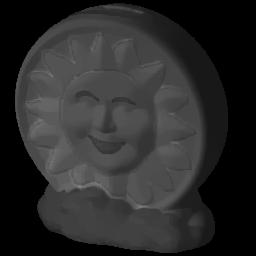}   & 
		\includegraphics[width=0.11\textwidth]{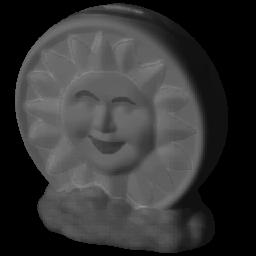}  \\		
        \includegraphics[width=0.11\textwidth]{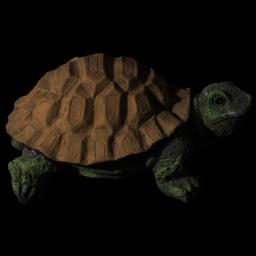} &
        \includegraphics[width=0.11\textwidth]{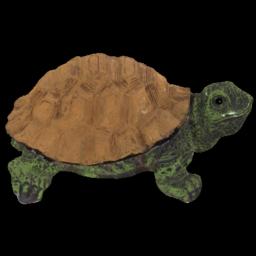}  &
        \includegraphics[width=0.11\textwidth]{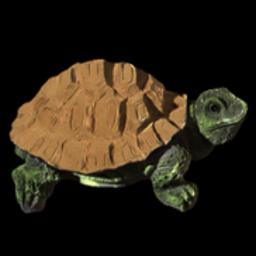}  &
        \includegraphics[width=0.11\textwidth]{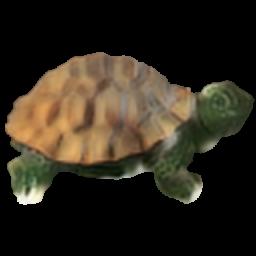}   & 
        \includegraphics[width=0.11\textwidth]{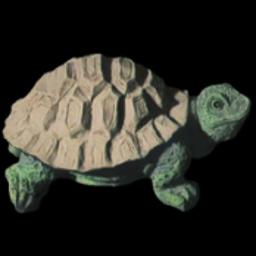}  &          
		\includegraphics[width=0.11\textwidth]{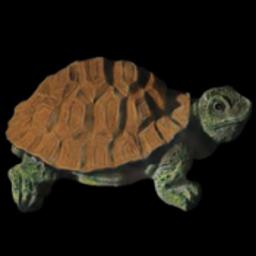}  &
\includegraphics[width=0.11\textwidth]{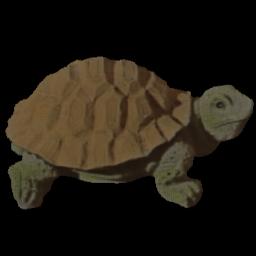}  & 
		\includegraphics[width=0.11\textwidth]{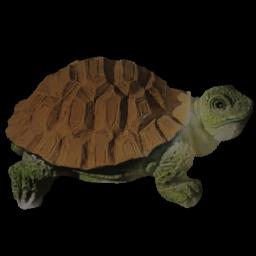}  \\		
        \includegraphics[width=0.11\textwidth]{figs/MIT/ones.png}  &
        \includegraphics[width=0.11\textwidth]{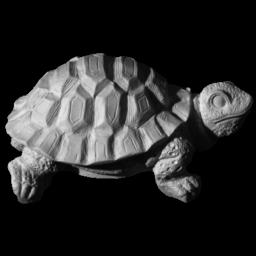}  &
        \includegraphics[width=0.11\textwidth]{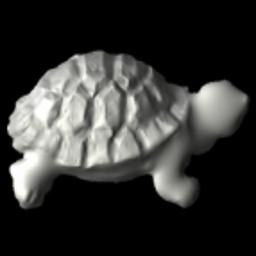}  &
        \includegraphics[width=0.11\textwidth]{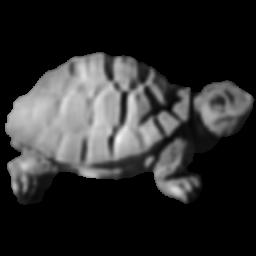}  & 
        \includegraphics[width=0.11\textwidth]{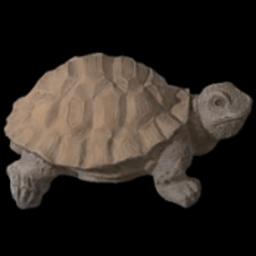}  &         
		\includegraphics[width=0.11\textwidth]{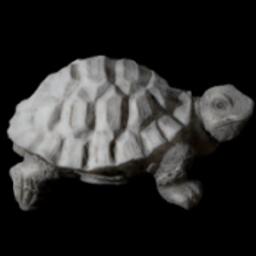}  &
		\includegraphics[width=0.11\textwidth]{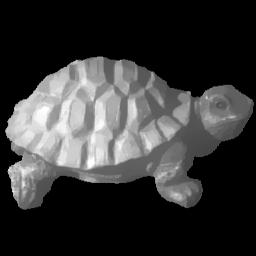}  & 
		\includegraphics[width=0.11\textwidth]{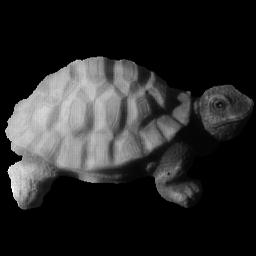}  \\						
        {\scriptsize Image} & {\scriptsize GT} & {\scriptsize SIRFS~\cite{barron2015shape}} & {\scriptsize DI~\cite{narihira2015direct}} & {\scriptsize Shi~\etal~\cite{shi2016learning}}  & {\scriptsize Shi~\etal~\cite{shi2016learning}$^{\star}$} & {\scriptsize Ours} & {\scriptsize $\text{Ours}^{\star}$} 

    \end{tabular}  
  \caption{ \textbf{Qualitative comparisons on MIT intrinsics
      testset.} Odd rows: reflectance predictions. Even rows: shading
    predictions. ${}^{\star}$ are the predictions fine-tuned on MIT.} \label{fig:MIT_comparisons}
\end{figure}

For the sake of completeness, we also test the ability of our
\ICGShort-trained networks to generalize to the MIT Intrinsic Images
dataset~\cite{grosse2009ground}.  In contrast to IIW/SAW, the MIT
dataset contains 20 real objects with 11 different illumination
conditions. We follow the same train/test split as
Barron~\etal~\cite{barron2015shape}, and, as in the work of
Shi~\etal~\cite{shi2016learning}, we directly apply our \ICGShort
trained networks to MIT testset, and additionally test fine-tuning them
on the MIT training set.


We compare our models with several state-of-the-art learning-based
methods using the same error metrics
as~\cite{shi2016learning}. Table~\ref{tb:tb_MIT} shows quantitative
comparisons and Figure~\ref{fig:MIT_comparisons} shows qualitative
results. Both show that our \ICGShort-trained model yields better
performance compared to ShapeNet-trained networks both qualitatively
and quantitatively, even though like MIT, ShapeNet consists of images
of rendered objects, while our dataset contains images of
scenes. Moreover, our \ICGShort-pretrained model also performs better
than networks pretrained on ShapeNet and Sintel. These results further
demonstrate the improved generalization ability of our \ICGShort
dataset compared to existing datasets. Note that SIRFS still achieves
the best results, but as described
in~\cite{narihira2015direct,shi2016learning}, their methods are
designed specifically for single objects and generalize poorly to real
scenes.

\section{Conclusion}

We presented a new synthetic dataset for learning intrinsic images,
and an end-to-end learning approach that learns better intrinsic image
decompositions by leveraging datasets with different types of
labels. Our evaluations illustrate the surprising effectiveness of our
synthetic dataset on Internet photos of real-world scenes. We find
that the details of rendering matter, and hypothesize that improved
physically-based rendering may benefit other vision tasks, such as normal prediction and semantic segmentation~\cite{zhang2017physically}.

\smallskip
{\small
\noindent \textbf{Acknowledgments.} We thank Jingguang Zhou for his help with
data generation. This work was funded by the National
Science Foundation through grant IIS-1149393, and by a grant from
Schmidt Sciences.
}

\clearpage

\bibliographystyle{splncs}
\bibliography{paper}

\end{document}